\documentclass[final,3p,times]{elsarticle}

\usepackage{hyperref}
\usepackage{lineno}
\modulolinenumbers[1]

\journal{Journal of \LaTeX\ Templates}

\usepackage{graphicx}
\DeclareGraphicsExtensions{.jpeg, .png, .jpg, .bmp, .tif, .PNG, .eps}
\usepackage[caption=false]{subfig}
\usepackage{amsmath} 
\usepackage{amsfonts}
\usepackage{amssymb}
\usepackage{bm}
\usepackage{algorithm}
\usepackage{algorithmic}
\usepackage{gensymb} 
\usepackage{array}
\usepackage{xcolor}

\usepackage{booktabs}
\usepackage{multirow}
\usepackage{multicol}
\usepackage{tabularx}
\makeatletter
\newcommand{\Bhline}{%
    \noalign {\ifnum 0=`}\fi \hrule height 1pt
    \futurelet \reserved@a \@xhline
}

\begin{document}

\begin{frontmatter}

\title{Drosophila-Inspired 3D Moving Object Detection Based on Point Clouds}

\author[NUDT]{Li Wang}\ead{wangli2014@nudt.edu.cn}
\author[NIIDT]{Dawei Zhao}

\author[NUDT]{Tao Wu}
\author[NUDT]{Hao Fu}
\author[NUDT]{Zhiyu Wang}
\author[NIIDT]{Liang Xiao}

\author[NUDT]{Xin Xu}
\author[NUDT,NIIDT]{Bin Dai\corref{mycorrespondingauthor}}
\cortext[mycorrespondingauthor]{Corresponding author}\ead{bindai.cs@gmail.com}

\address[NUDT]{College of Intelligence Science and Technology,\\ 
 National University of Defense Technology, Changsha, China}
\address[NIIDT]{Unmanned Systems Research Center,\\
 National Innovation Institute of Defense Technology, Beijing, China}

\begin{abstract}
3D moving object detection is one of the most critical tasks in dynamic scene analysis. In this paper, we propose a novel Drosophila-inspired 3D moving object detection method using Lidar sensors. According to the theory of elementary motion detector, we have developed a motion detector based on the shallow visual neural pathway of Drosophila. This detector is sensitive to the movement of objects and can well suppress background noise. Designing neural circuits with different connection modes, the approach searches for motion areas in a coarse-to-fine fashion and extracts point clouds of each motion area to form moving object proposals. An improved 3D object detection network is then used to estimate the point clouds of each proposal and efficiently generates the 3D bounding boxes and the object categories. We evaluate the proposed approach on the widely-used KITTI benchmark, and state-of-the-art performance was obtained by using the proposed approach on the task of motion detection.

\end{abstract}


\end{frontmatter}

\section{Introduction}\label{sec:introduction}

Dynamic scene understanding is a major challenge for autonomous vehicles or mobile robots, where the detection of moving objects is a fundamental task. Moving objects, such as walking pedestrians or moving cars, pose serious threats to autonomous vehicles and directly affect the path planning results. In addition, the detection of moving objects is one of the main difficulties for the localization and mapping of autonomous vehicles. Existing simultaneous localization and mapping (SLAM) and visual odometry (VO) techniques usually treat the moving objects as outliers, so robust algorithms need to be designed to weaken the influence of those moving objects. On the other hand, the dynamic objects contain valuable motion information, which tends to benefit road layouts prediction and scene modeling.

For dynamic object detection, both 2D and 3D detection approaches could be adopted. Since a 2D image is a projection of the 3D world space, it can hardly recover the 3D information of the scene. Therefore, 3D object detection approaches are usually adopted for autonomous vehicles. Stereo cameras and Lidars are the two most commonly used 3D sensors for autonomous vehicles. Although enabling rich color, texture and 3D information, stereo cameras are sensitive to luminance changes which make it impossible for high-precision 3D information. In contrast, Lidars can accurately capture 3D information in real-time. This paper aims to develop a simple and efficient 3D moving object detection approach using Lidar sensor. 

In recent years,  object detection technology has developed by leaps and bounds, which has accordingly promoted moving object detection research. Existing moving object detection approaches in 3D scenes can be roughly divided into two categories. The first category method is based on a combination of object detection and tracking \cite{Himmelsbach2012Tracking,luo2018fast}. It first detects all potential objects in the scene and then tracks them to judge whether they are moving or not. Although this type of method is simple in idea and easy to implement, its performance is not robust enough because the two subtasks of object detection and multi-object tracking have their own challenges, especially the imperfect accuracy and stability of multi-object tracking \cite{luo2014multiple}. The second category method attempts to skip the object detection part, and directly tracks each point or pixel using optical/scene flow approaches \cite{Lenz2011Sparse,Long2017Moving,Xiao2017Dense}, which becomes the mainstream method for moving object detection. Based on the tracked points, the ego-motion is firstly estimated and then compensated to estimate the real motion of each point in the world space. However, this kind of method is susceptible to background motion noise and not sensitive to motion. Furthermore, the bottom-to-up optical/scene flow calculation also causes expensive computational cost. Therefore, motion detection is still an unresolved problem.

In nature, Drosophila is an insect that is particularly sensitive to motion, making it easy to escape the predation of enemies. Some research tries to reveal the mechanism of motion detection for these insects. Among them, the elementary motion detector (EMD) theory \cite{hassenstein1956system} is a promising approach. Unlike the detection-tracking and optical/scene flow method, this theory realizes motion detection by simulating the Drosophila visual circuits. Some previous works try to apply EMD to surveillance camera-based 2D object detection \cite{Pallus2010Modeling,wang2018directionally}. In this paper, we extend the EMD idea to the more complex 3D Lidar point clouds and develop a simple and efficient motion detection approach.

\begin{figure*}[t]
	\centering
	\includegraphics[scale=0.39]{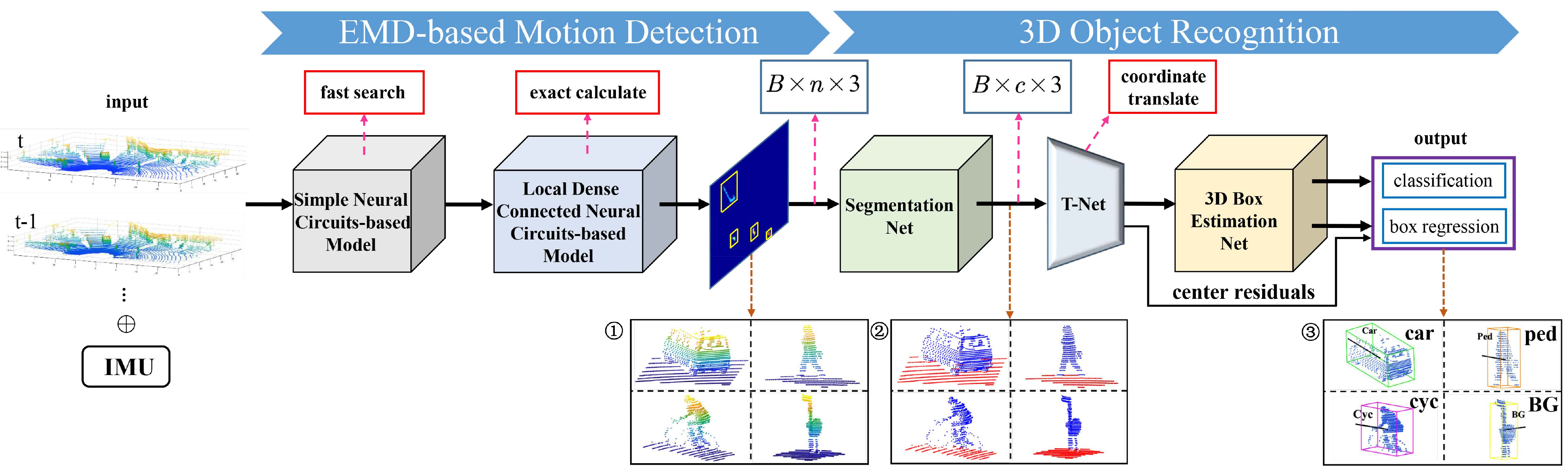}
	\caption{\textbf{The pipeline of the Drosophila-inspired 3D moving object detection.} Taken lidar point cloud sequences and IMU as input, an EMD-based motion detector calculates all motion areas of the scene and extracts the moving raw point clouds (details in Sec.\ref{sec:EMD}). Then the segmentation net labels the foreground points. The 3D box estimation network classifie objects and regresses 3D box parameters (details in Sec.\ref{sec:3Ddetection}). \textcircled{1} $\sim$ \textcircled{3} express the results of current step.}
	\label{architecture}
\end{figure*}

In this work, we present a novel 3D moving object detection approach called EMD-PointNet. As illustrated in Fig.\ref{architecture}, given consecutive Lidar point clouds as input, the proposed approach can accurately detect the moving objects in 3D scenes. Our approach primarily consists of two stages: In the first stage, the EMD-based motion detector searches for motion areas and calculates motion information in a coarse-to-fine fashion to form moving object proposals. In the second stage, the point clouds of the proposals are segmented into foreground and background, and the foreground points are then classified and regressed to generate the 3D bounding boxes by neural networks. Our 3D detection network is based on the Frustum-PointNet \cite{qi2018frustum}, but we reduce the size of the network to one-third of the original one and improve the calculation of 3D box regression parameters. The proposed method is tested on the KITTI benchmark \cite{Geiger2012CVPR}. Compared with the start-of-the-art methods, our approach shows competitive results.

In summary, the main contributions of this paper are as follows:
\begin{itemize}
\item Based on the theory of elementary motion detector, we propose a novel and efficient motion detection approach for Lidar point clouds. To the authors' best knowledge, this paper is the first work to expand the Drosophila visual model to 3D space.
\item We improve a state-of-the-art 3D object detection network by optimizing the solution method and simplifying the model to achieve better performance and high efficiency.
\item We perform extensive experiments and obtain state-of-the-art results on a publicly available dataset.
\end{itemize}

The remainder of this paper is organized as follows: In Section \ref{sec:relatedworks}, we review the related work. Section \ref{sec:Framework} shows the entire framework of our approach. Section \ref{sec:EMD} and Section \ref{sec:3Ddetection} introduce the bio-inspired motion detection algorithm and the 3D object detection network. The qualitative and quantitative experiments on the KITTI benchmark are described in Section \ref{sec:Experiments}, followed by the conclusions in Section \ref{sec:conclusion}.

\section{Related Work}\label{sec:relatedworks}
Research topics related to our work include motion detection, 3D object detection, and elementary motion detector (EMD). In this section, we briefly review the existing works, especially those focusing on autonomous driving.

\textbf{Motion Detection.}
As a fundamental technology in autonomous driving, motion detection has been extensively studied. Various approaches have been developed for different data input, including image sequences, stereo sequences, 3D point clouds, etc.

For image sequences, the classical interframe difference methods \cite{xiaoyang2013novel,lee2015moving} and the background differential methods \cite{zivkovic2006efficient,chen2019background} are widely used for stationary cameras. The optical-flow-based methods are usually used for complex dynamic backgrounds \cite{Guizilini2013Online, Nguyen2015Optical}. Some recent works \cite{sevilla2016optical,SMSnet2017,MODNet2018} combine semantic information with optical flow to improve the accuracy of motion detection. However, as the moving object detection results are presented on the 2D image coordinate system, these methods play a limited role in mobile robots.

Compared with image sequences, stereo images provide not only color and texture but also the depth information. The moving objects could be detected in 3D space using scene flow approaches \cite{Lenz2011Sparse,Menze2015Object,Taniai2017Fast}. In \cite{Giosan2014Superpixel}, a coarse-to-fine method that combines RGB, depth and optical flow information was proposed to detect moving objects. Chen et al. \cite{Long2017Moving} utilized the disparity map and a modified random sample consensus (RANSAC) algorithm to detect the dynamic points. The moving objects were then extracted by a clustering algorithm. Zhou \cite{Zhou2017Moving} calculated the motion likelihood for each pixel by modeling the ego-motion uncertainty, and used the Graph-Cut algorithm to segment moving objects. To reduce the depth uncertainty from stereo vision, Xiao et al. \cite{Xiao2017Dense} constructed a unified conditional random field (CRF) model and decomposed the dynamic region segmentation into two separate stages. ISF \cite{Behl2017Bounding} embedded 2D instance segmentation into a CRF model to accurately estimate 3D scene flow and detect dynamic objects. The methods above generally use flow information for motion detection and segmentation. Due to the susceptibility to background motion noise, these methods are not robust and stable enough. In addition, the clustering algorithm is also used to detect moving areas. Sandhu et al. \cite{sandhu2018fast} presented a spectral graph clustering method to segment dynamic objects. Kochanov et al. \cite{kochanov2016scene} built semantic 3D occupancy grid maps and adopted density-based spatial clustering to detect objects. However, these methods may fail in spatial clustering due to many constraints. Despite the remarkable achievements in motion detection on the basis of stereo pairs, its accuracy is still not comparable with Lidar-based approaches owing to the limited accuracy of the stereo depth estimation.

For Lidar point clouds, early works address this problem through detection and multi-frame tracking. In \cite{Himmelsbach2012Tracking}, the top-down knowledge about the tracked objects was utilized for improving the dynamic segmentation performance. DATMO \cite{asvadi2015detection} generated a 2.5D map based on perception and localization data, and used spatio-temporal information to detect and track dynamic objects. Based on the likelihood-field vehicle model, LFMDT \cite{Chen2016Likelihood} detected and tracked moving vehicles using a modified Scaling Series algorithm. However, complex models and unstable tracking make these methods inefficient.
In order to improve efficiency, the Bayesian occupancy filter (BOF) is presented to detect dynamic objects. Without a complete SLAM process, a BoF-based method used odometer/IMU to transmit occupancy information and detected moving objects \cite{baig2012fast}. Nuss et al. \cite{nuss2015fusion} used sequential Monte Carlo BOF to detect motion at a low semantic level. Nevertheless, these methods are not precise enough owing to the inaccuracy of Bayesian occupancy model. In addition, deep neural networks and point-flow have become popular for lidar-based moving object detection. In \cite{luo2018fast}, a deep network was proposed to jointly detect, track and predict the motion of the objects from lidar sequences represented in the bird's eye view. Ushani et al. \cite{ushani2017learning} estimated scene flow using an expectation-maximization algorithm in occupancy grids. Based on the assumption of local geometric constancy, an energy function was formulated to estimate the Lidar scene flow in \cite{dewan2016rigid}, which added local rigid regularization to smooth motion field. Recently, a novel deep neural network named FlowNet3D \cite{liu2019flownet3d} was proposed. It estimated the point flow between consecutive frames in an end-to-end fashion. PointFlowNet \cite{behl2019pointflownet} is another deep neural network that jointly estimates the point-based scene flow and the ego-motion. A drawback of these approaches is the high computational costs. Unlike the methods above, our approach efficiently detects motion areas through an improved biological model instead of probability and neural network models.

\textbf{3D Object Detection.}
In the realm of mobile robotics, particularly autonomous driving, 3D object detection is a fundamental task. There exists a large amount of literature on related works. In this paper, we focus on those works that process lidar point clouds using deep neural networks.

As a milestone approach, PointNet \cite{qi2017pointnet} and PointNet++ \cite{qi2017pointnet++} are the first to design a network that can directly process the unordered raw point clouds. They can classify and segment point clouds with high precision by learning the local and global features of point clouds. Based on these works, Qi \cite{qi2018frustum} projected 2D object boxes into 3D space to form frustum points and then regressed 3D boxes. In contrast, Lidar points voxelization is another processing method. VoxelNet \cite{zhou2018voxelnet} divided point clouds into voxels and extracted local features of each voxel using PointNet. It then used the region proposal network to detect objects. PointPillars \cite{lang2018pointpillars} didn't divide vertical columns into fine-grained voxels. Each vertical column was treated as a whole and processed as one pillar. The features for each pillar were then extracted through PointNet. Recently, using 2D projection view to assist the processing of 3D point cloud has become popular. MV3D \cite{chen2017multi} fused the frontal view as well as the bird's eye view (BEV) projection to detect 3D objects. AVOD \cite{ku2018joint} designed a 3D-RPN-based network architecture based on images and lidar BEV maps, and obtained a better result. PIXOR \cite{yang2018pixor} treated the BEV map of height and reflectivity as the input and used the RetinaNet for object detection and localization. Although the voxel-based methods and BEV-based methods are simple to process, they also lose some features of the original point clouds. Therefore, the approach proposed in this paper designs an efficient network to directly process point clouds.

\textbf{Elementary Motion Detector.}
The seminal work on EMD was proposed in \cite{hassenstein1956system}. Since then, the neurobiologists continue to advance the research on elementary motion detectors in recent decades. Pallus et al. \cite{Pallus2010Modeling} proposed an architecture composed of a grid of correlation-type EMD. In \cite{eichner2011internal,maisak2013directional}, researchers found that Drosophila divides the input signal into two parallel channels according to brightness change. Compared with CNN, there are few engineering studies and applications about EMD. Considering motion direction and velocity estimation, Wang \cite{wang2016bio} presented a lateral inhibition model to improve the performance of small object detection. In \cite{wang2018directionally}, a directionally selective EMD was used to detect small dynamic objects. In addition, Wang improved the model for small dynamic objects, including setting up feedback as well as fusing the multi-field visual features \cite{wang2017improved,wang2018feedback}. Even though these methods that simply used EMD for images still have some problems such as object deformation, the thought of EMD has greatly inspired moving object detection based on Lidar points in this paper.

\section{Framework}\label{sec:Framework}

Our approach aims to efficiently detect the moving objects in 3D scenes for autonomous driving. To this end, we develop a novel 3D motion detection method that is inspired by the theory of Elementary Motion Detector from insect compound eye structures. As shown in Fig. \ref{architecture}, the proposed approach combines the shallow visual neural pathway of Drosophila with a brain-inspired cognitive neural network to implement motion detection and object recognition, respectively. It can be divided into two stages: motion detection and object recognition. Taking point cloud sequences and inertial measurement unit (IMU) data as the input, the EMD-based motion detector fast calculates the motion information in a coarse-to-fine fashion and extracts point clouds of each motion area to form dynamic object proposals. The object detection network then labels the foreground point clouds, and estimates them to generate the 3D bounding boxes and object categories. This approach will be introduced in Sec. \ref{sec:EMD} and Sec. \ref{sec:3Ddetection}. 

It is acknowledged that motion and stationary are relative concepts in physics. The determination of motion depen- ds on the choice of the reference system. In this work, the Earth coordinate system is chosen as the reference system.

\section{EMD-based Motion Detection}\label{sec:EMD}
In this section, we first briefly introduce the elementary motion detector theory, which is the theoretical basis of our motion detection method. Then we will describe our motion detection method, including the preprocessing, motion detection, and multi-frame fusion.

\subsection{Elementary Motion Detector Theory}\label{sec:EMD_theory}

\begin{figure}[!t]
	\centering
	\includegraphics[scale=0.38]{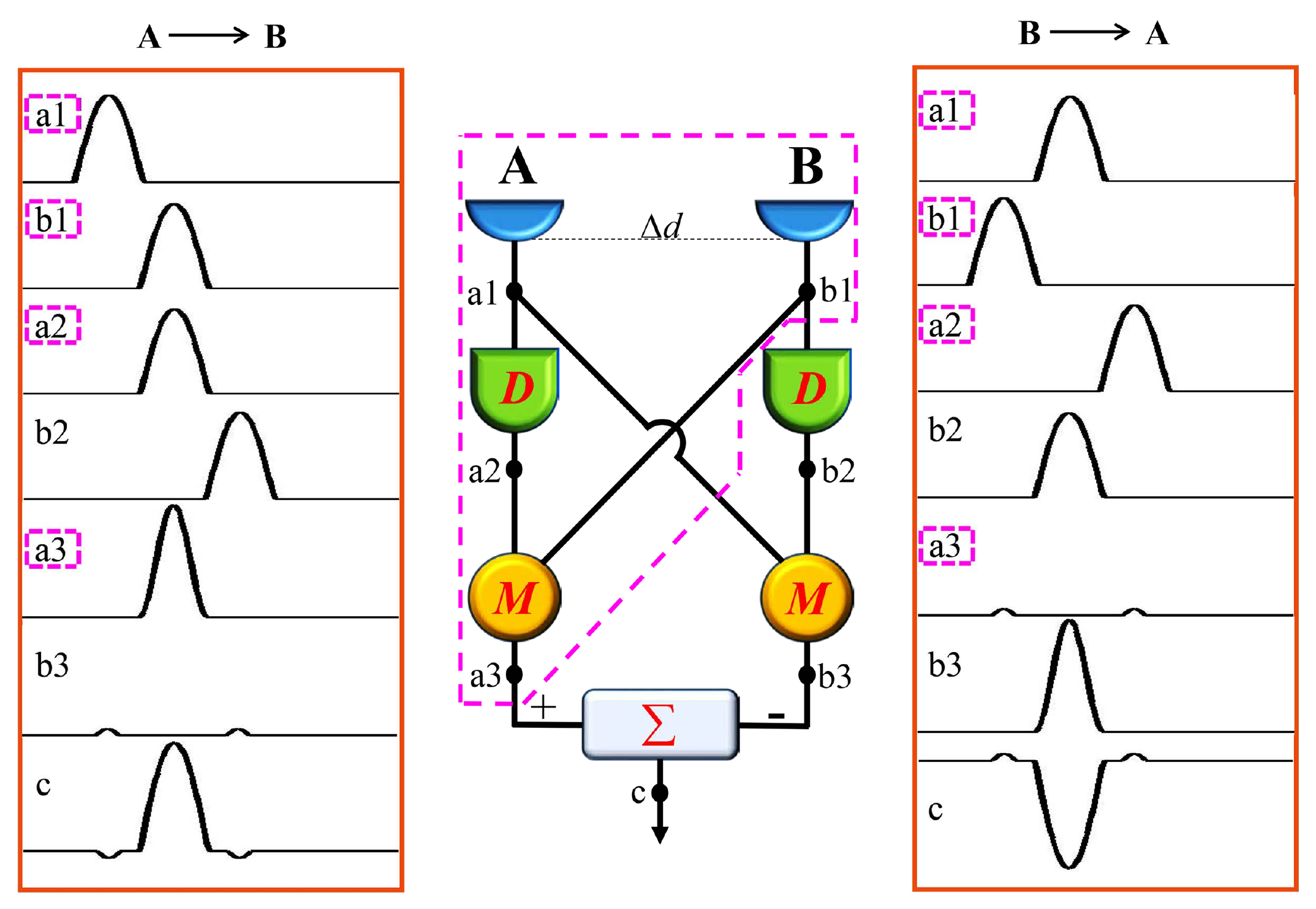}
	\caption{\textbf{EMD Theoretical Model.} This structure is a standard EMD model, where the blue semicircle denotes the receptor, which is equivalent to each small eye in the compound eye; the green part is a delay unit; the orange circle represents the multiplier module, and the light grey component is the summation unit. The magenta dotted line part represents a single-arm EMD model. The left waveforms describe the movement from A to B, and the opposite is shown on the right (details in Sec. \ref{sec:EMD_theory}).}
	\label{EMD_model}
\end{figure}

We take the single-arm EMD model as an example to introduce the principle, as shown in the magenta dashed line of Fig. \ref{EMD_model}. The elementary motion detector model consists of a pair of adjacent receptor channels with a baseline distance of $\Delta d$. Suppose a moving light spot sequentially passes through the adjacent receptors A and B. Receptor A channel delays the light signal for a certain time and makes it meet the signal in B channel. Multiplying these two signals will produce a large output response. Similarly, if the light spot moves from receptor B to A, the signals cannot meet after the model processing. So the output response is 0 in the ideal case. In practice, it's usually a tiny value. In this way, the two motion directions are clearly separated. 

In order to improve the specificity of direction selection, EMD usually adds another pair of mirror-symmetrical channels, then subtracts the two channels to obtain the final result. In this way, the responses of two opposite motions are equal in magnitude and opposite in sign. The architecture is shown in Fig. \ref{EMD_model}.

\subsection{Preprocessing}
Inspired by EMD, we extend the model for Lidar point cloud processing. Since the Lidar points are usually disordered compared to the regularly arranged small eyes in the compound eye, we have to map the raw Lidar points into voxels for further processing. While voxels can be regarded as receptors of the EMD model, conventional fine-grained voxels \cite{zhou2018voxelnet} are not suitable for the EMD model. Because the low vertical resolution makes it difficult to find one-to-one corresponding fine-grained voxels in the dynamic Lidar sequence space. Decreasing the vertical resolution of voxels can increase the shape stability of the object. Therefore, unlike fine-grained voxelization, the voxel resolution is set to $0.2\times0.2\times \Delta h$, where $\Delta h$ is determined by the highest and lowest points in the voxel. The bird's eye view maps with two channels are then generated. One channel is a binarized channel indicating whether the corresponding voxel is occupied. It is set to 1 if at least one point falls in the voxel. The other channel encodes the average height of point clouds in a voxel. This representation can maximize the usage of stable object shapes from BEV maps and increase computational efficiency. Note that the ground points are removed by the RBNN algorithm \cite{Klasing2008RBNN} before point clouds voxelization. In addition, the IMU data is utilized to calculate the vehicle's ego-motion. The previous data are projected to the current Lidar coordinate system according to the ego-motion. Let $P_{t-1} = \left[ x\ y\ z\ 1 \right] ^T$ represent a 3D point in the Lidar coordinate system at t-1, it is projected to the Lidar coordinate system at time t by

\begin{equation}
\begin{aligned}
P_{t-1}^{'} \, = \,\textbf{T}_{t}^{-1} \cdot \textbf{T}_{t-1} \cdot P_{t-1} \, ,
\end{aligned}
\label{coord trans}	
\end{equation}
where $\textbf{T}_{t-1}$ and $\textbf{T}_{t}$ are the transformation matrix from the Lidar coordinate system to the global coordinate system at time t-1 and t respectively.   

\subsection{Motion Detection}\label{sec:EMD_main}
Utilizing the detection characteristic of optional motion direction, the proposed motion detection algorithm adopts a coarse-to-fine strategy to improve the computational efficiency. It is divided into two stages: a fast search stage and an exact match stage. In the first stage, our algorithm uses simple neural circuits connections to roughly calculate the approximate motion direction of the object; in the second stage, our algorithm exactly matches objects in a local range to obtain the accurate motion information by dense neural circuits connections. The details of this algorithm are as follows.

\begin{figure}[!t]
	\centering
	\includegraphics[scale=0.32]{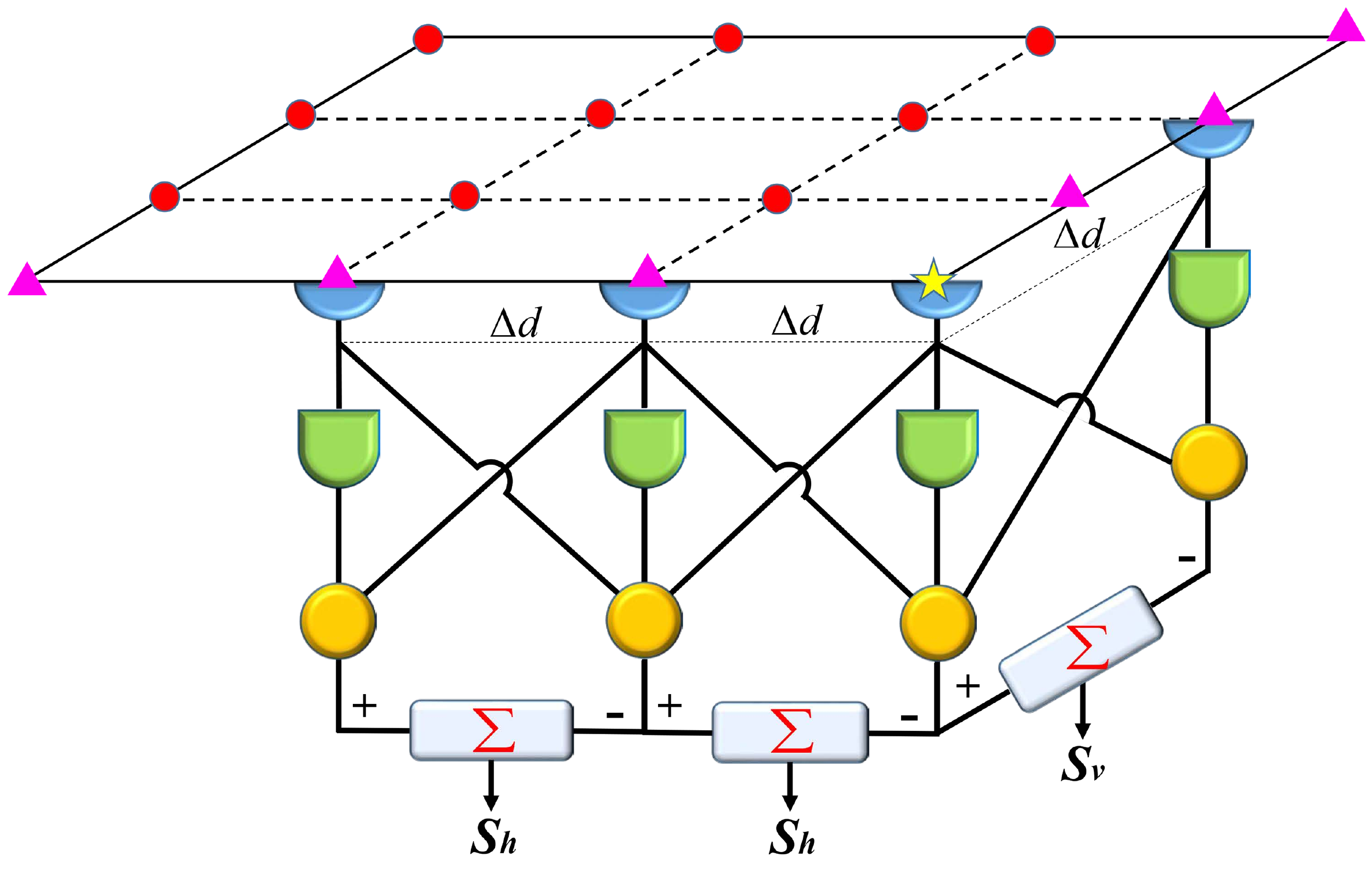}
	\caption{\textbf{EMD Connection Model.} Simple connection means the receptors are connected in horizontal and vertical directions, i.e. the pentagram is only connected with triangles. Dense connection means omnidirectional connection, i.e. the pentagram is connected with triangles and circles. Changing the distance of interconnected receptors, this model could detect the moving objects with different velocities.}
	\label{2DMD}
\end{figure}

For the fast search stage, the standard EMD model is extended to the 2D BEV occupation map to detect the movement tendency, which is established only on horizontal and vertical cells, as shown in Fig. \ref{2DMD}. The input BEV occupation maps are filtered by a first-order low-pass filter to mimic the blur of fly optics and slight signal delay, where the filter satisfies Eq. \eqref{first-order LPF}:

\begin{equation}
\begin{aligned}
\frac{dI^{f}\left( t \right)}{dt} \, = \,\frac{1}{\tau}\left( I\left( t \right) -I^{f}\left( t \right) \right) \, ,
\end{aligned}
\label{first-order LPF}	
\end{equation}
where $\tau$ is the time constant, $I\left( t \right)$ is the current BEV map, and $I^{f}\left( t \right)$ is the filtered output, corresponding to a2 and b2 in Fig. \ref{EMD_model}. The filter will create a motion blur effect, which reflects the movement tendency.

In order to adapt to different velocities, various templates are used to detect motion by changing the connection distance of the receptors. The maximum connection distance $ R $ determines the upper-bound detection velocity. In a search area with the size of $2R \times 2R$, the horizontal and vertical motion scores satisfy the following equations:
\begin{eqnarray}
\label{EMD_velocity1}
&S_{h \, i,j}\left[ x \right] = I^{f}_{i,j}\left( t \right) \cdot I_{i+x,j}\left( t \right) - I_{i,j}\left( t \right) \cdot I^{f}_{i+x,j}\left( t \right) \, , \\
\label{EMD_velocity2}
&S_{v \, i,j}\left[ y \right] = I^{f}_{i,j}\left( t \right) \cdot I_{i,j+y}\left( t \right) - I_{i,j}\left( t \right) \cdot I^{f}_{i,j+y}\left( t \right) \, ,
\end{eqnarray} 
where $x,y = \{ a | a \in \left[ -R,R\right], a\in \mathbb{Z} \}$, which represent the local coordinates in the search area; $I^{f}_{i,j}\left( t \right)$ is the delayed signal at $\left( i,j \right)$ cell; $I_{i,j}\left( t \right)$ is the current signal. The maximum of $S_h$ and $S_v$ can reflect the approximate motion direction of each cell, and the $x$ and $y$ values corresponding to the maximum motion score are recorded as $x_{sm}$ and $y_{sm}$. Then, the motion direction vector can be expressed as $\left(x_{sm},y_{sm}\right)$. At the same time, the moving and non-moving areas are also determined in the fast search stage. 

Based on the first stage results, a local template matching algorithm for the moving area is proposed to accurately calculate the motion vectors of each cell in the exact match stage. Unlike cell-to-cell calculation in the fast search stage, each moving cell is extended to an $m\times m$ patch as the operation unit in this stage. For each moving cell $\left(i,j\right)$ in the BEV map, a local search space $\mathbb{S}$ is defined as a right-angled sector area with $\left(x_{sm},y_{sm}\right)$ as the angular bisector and $R$ as the radius, where the moving cell is the origin of this space and the local coordinates of other cells in this space are written as $\left(x,y\right)$. Unlike the simple connection shown in Fig. \ref{2DMD}, a fully neural circuits connected EMD model is developed to create more delicate speed templates that can accurately calculate motion information in the search space. The input signals include BEV height map $I^{h}$, BEV occupancy map $I$, and Gaussian filtered map $I^{g}$. When the BEV height map and BEV occupancy map are taken as the input, the motion detection model changes subtraction operators instead of multiplication operators to better adapt Lidar data. For three input signals, each cell $\left(i,j\right)$ creates three energy functions to represent the motion response, as follows: 
\begin{eqnarray}
\label{E1}
E_{1}\left( x,y \right) &= &\sum_{a=1}^m \sum_{b=1}^m \left|I_{a,b}^{g}\left( t \right) \cdot I_{a+x,b+y}^{g}\left( t-1 \right)\right|  \, , \\  
\label{E2}
E_{2}\left( x,y \right) &= &\sum_{a=1}^m \sum_{b=1}^m \left|I_{a,b}\left( t \right) - I_{a+x,b+y}\left( t-1 \right)\right|  \, , \\  
\label{E3}
E_{3}\left( x,y \right) &= &\sum_{a=1}^m \sum_{b=1}^m \left|I_{a,b}^{h}\left( t \right) - I_{a+x,b+y}^{h}\left( t-1 \right)\right| \, ,
\end{eqnarray} 
where $E_{1}\left( x,y \right)$ denotes the motion score of the Gaussian filtered map. A higher score indicates a better local matching result. $E_{2}\left( x,y \right)$ indicates the similarity of the two patches. If two patches are quite similar, the value would be close to 0. In the same way, $E_{3}\left( x,y \right)$ expresses the similarity of height distribution. To fuse these information, the local template matching algorithm is formulated by a total energy function:

\begin{equation}
\label{Energy}
E\left( x,y \right) = \, \bm{\omega_{1}}\left(1-E^{'}_{1}\left(x,y\right)\right) + \, \bm{\omega_{2}}E^{'}_{2}\left(x,y\right) + \, \bm{\omega_{3}}E^{'}_{3}\left(x,y\right) \, ,	
\end{equation}
\begin{equation}
\label{E_min}
\bm{\upsilon}_{i,j} = \left( x^{'},y^{'}\right) = \underset{x,y \, \in \, \mathbb{S}}{arg \min} \, E\left( x,y \right) \, ,
\end{equation}
where $\bm{\omega}$ is the weight, and $E^{'}$ is the normalized energy, which denotes the motion score of different type input signals. The motion vector of a cell, $\bm{\upsilon}_{i,j}$, can be calculated by minimizing the energy function.

This coarse-to-fine strategy reduces the computational complexity to about 1/4 of the exhaustive search without losing the detection accuracy.

In order to reduce the impact of background noise and IMU errors, a lateral inhibition method similar to \cite{wang2016bio} is used to filter motion vectors. The motion vectors can be divided into horizontal components $\bm{v}_{x}$ and vertical components $\bm{v}_{y}$, written as:

\begin{equation}
\begin{aligned}
\bm{v}_{x} = \left[ x^{'} \left( i,j \right) \right]_{M \times N} \, , \, \,
\bm{v}_{y} = \left[ y^{'} \left( i,j \right) \right]_{M \times N} \, ,
\end{aligned}
\label{velocity}	
\end{equation}
where $x^{'} \left( i,j \right)$ and $y^{'} \left( i,j \right)$ represent the motion vectors calculated in Eq.\eqref{E_min}; $M$ and $N$ are the dimensions of the grid map. A filter kernel $F$ convolutes these two motion matrices, which is defined as:

\begin{eqnarray}
\label{Filter}
F = \left( 
\begin{matrix}
	q&		q&		\cdots&		q&		q\\
	q&		0&		\cdots&		0&		q\\
	\vdots&		\vdots&		p&		\vdots&		\vdots\\
	q&		0&		\cdots&		0&		q\\
	q&		q&		\cdots&		q&		q\\
\end{matrix} 
\right)_{l\times l} , \, \, p + 4(l-1)q = 0  \, , \\
\label{Conv}
\bm{v}_{x}^{f} = \bm{v}_{x} \otimes F \, , \hspace{2.4cm}  \\
\bm{v}_{y}^{f} = \bm{v}_{y} \otimes F \, , \hspace{2.4cm}
\end{eqnarray}
where $p$ and $q$ are the kernel weights; $\bm{v}_{x}^{f}$ and $\bm{v}_{y}^{f}$ are filtered motion results; $\otimes$ is convolution operator.
By choosing an appropriate filter size, this operator inhibits the motion consistency between the center and the surrounding areas but does not inhibit the real object movement. Therefore, it can reduce errors from large-area background movements or the whole scene motion caused by IMU zero drift.

The motion detection approach described in Algorithm \ref{A1:MotionDetector} could detect all moving areas in the scene. A major characteristic of this method is that the detector can detect the specific movements by changing the connection between the receptors according to the interested motion, thereby improving the calculation efficiency. For example, on the highway, the detector can efficiently detect vertical motion by only connecting the receptors in the vertical direction; when focusing on the opposite movement, the detector can use the single-arm EMD model to detect the unidirectional motion.

\begin{algorithm}[!t] 
\caption{ EMD-based Motion Detection } 
\label{A1:MotionDetector} 
\begin{algorithmic}[1] 
\REQUIRE Lidar sequences $P_{t}, P_{t-1}$; IMU transformation matrix $\bm{T_{t}}, \bm{T_{t-1}}$.
\ENSURE motion map $\bm{\upsilon} \left( t \right)$; moving object proposals $\mathbb{O}_{t}=\left\lbrace \bm{O}_{1}, \bm{O}_{2}, ... \right\rbrace $.
\STATE Unify Lidar sequences to the current Lidar coordinate system by Eq.\eqref{coord trans}; 
\STATE Voxelize Lidar points to form three BEV maps: height map $I^{h}$, occupancy map $I$ and Gaussian filtered map $I^{g}$;
\STATE First-order low-pass filter: $I^{f} = FoLPFilter(I)$;  
\WHILE {$i,j \in I^{f} $} 
\STATE Calculate horizontal and vertical motion scores by Eq.\eqref{EMD_velocity1} and Eq.\eqref{EMD_velocity2}: $S_{h \, i,j}\left[ x \right], S_{v \, i,j}\left[ y \right]$;
\STATE Rough motion vector: $\bm{v}^{r} \left( i,j \right) = \left( x_{sm}, y_{sm} \right) \gets max \left( S_{h \, i,j}\left[ x \right], S_{v \, i,j}\left[ y \right] \right)$;
\STATE Rough motion area: $I_{rm} \left( n \right) \gets \bm{v}^{r} \left( i,j \right) $ ;
\ENDWHILE  
\WHILE {$i,j \in I_{rm} $} 
\STATE Establish a dense local search space $\mathbb{S}$;
\STATE Calculate the matching scores by Eq.(\ref{E1} - \ref{E3}): $E_1, E_2, E_3$;
\STATE Exact motion vector: $\bm{\upsilon}_{i,j} \left( t \right) = arg \min \,  \left[ \bm{\omega_{1}}\left(1-E^{'}_{1}\right) + \, \bm{\omega_{2}}E^{'}_{2} + \, \bm{\omega_{3}}E^{'}_{3}\right] $;
\ENDWHILE
\STATE Cluster to form point-cloud-based moving object proposals: $\mathbb{O}_{t}=\left\lbrace \bm{O}_{1}, \bm{O}_{2}, ... \right\rbrace$;
\RETURN $\bm{\upsilon} \left( t \right), \mathbb{O}_{t}$. 
\end{algorithmic}
\end{algorithm}

\subsection{Multi-Frame Fusion}
The above algorithm can detect most of the moving objects in a scene. However, due to the limited resolution of the BEV map, some slow-moving objects still could not be detected, such as slow-moving pedestrians. Assuming that the sampling frequency of Lidar is 10 Hz, a pedestrian with a speed of 5 km/h moves only about 0.1 meters in 0.1 seconds. It's difficult to detect just using two consecutive frame data. In theory, the consecutive frames with a long time interval can help detect low-speed dynamic objects. Therefore, the proposed approach is improved by combining multiple frames with the current frame. Using a method similar to max-pooling, multi-frame results are fused to calculate the final motion scores.

Finally, the cells are clustered into individual objects based on the motion and geometry information. For each clustered object, the occupied voxels are slightly expanded, and all the points in these voxels are assembled to form a 3D moving object proposal for the next object recognition module.

\section{3D Object Detection Network}\label{sec:3Ddetection}
\begin{figure}[!t]
	\centering
	\subfloat[Segmentation Net]{\includegraphics[scale=0.34]{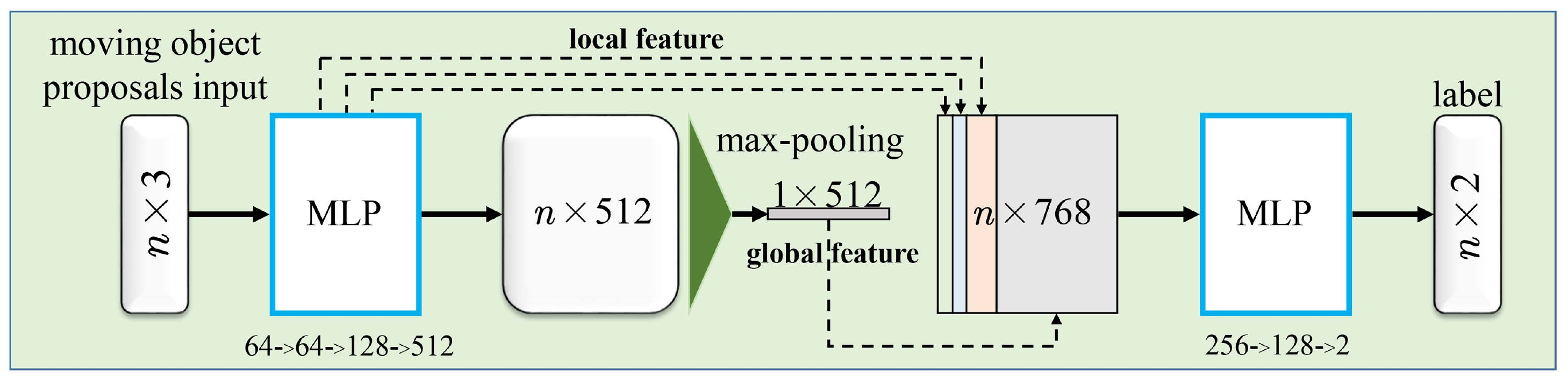}}\\
	\subfloat[T-Net]{\includegraphics[scale=0.34]{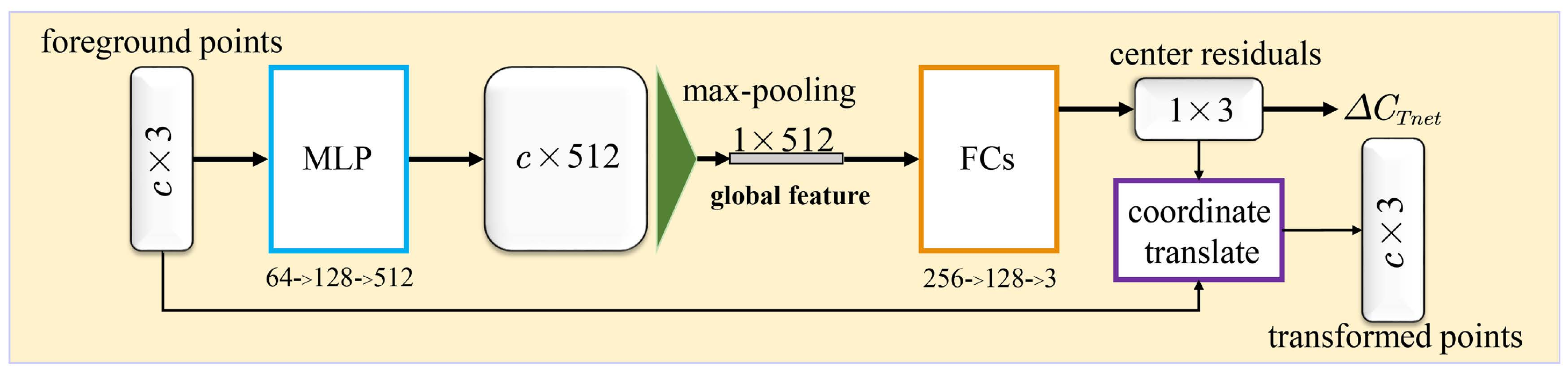}}\\
	\subfloat[3D Box Estimation Net]{\includegraphics[scale=0.34]{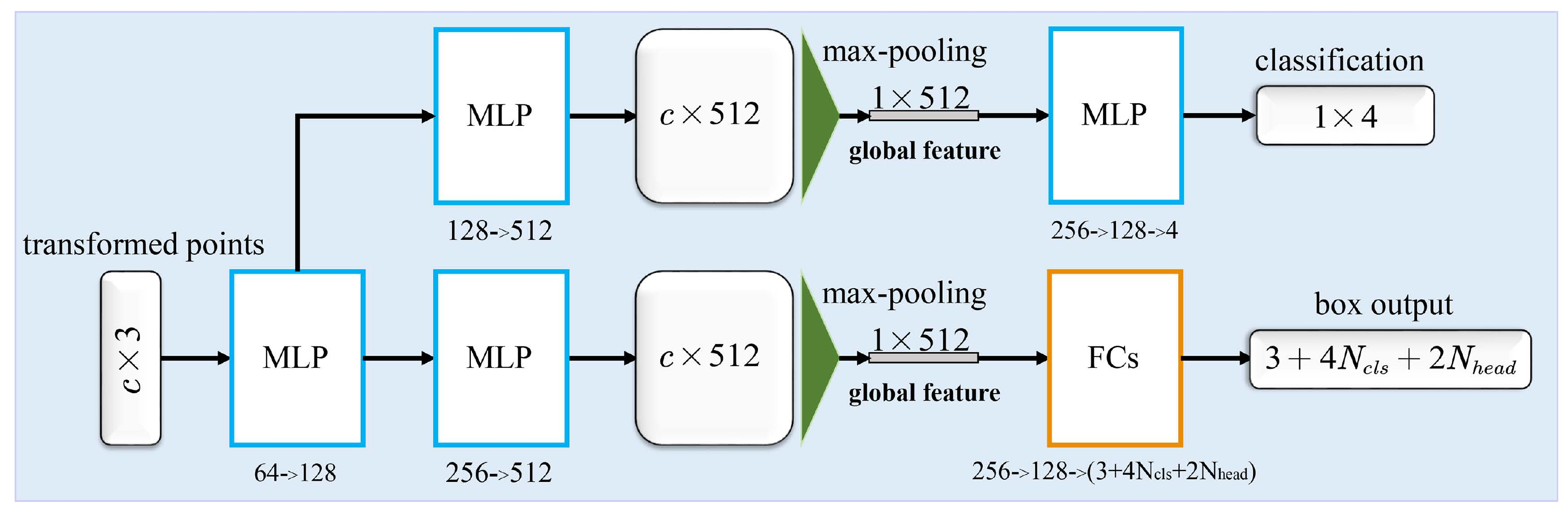}}\\
	\caption{\textbf{3D object detection network.} The detection network is cascaded by three sub-networks. The segmentation subnetwork (a) divides raw points of a proposal into foreground and background. The T-net (b) calculates the residuals of box center and transforms the coordinate of foreground points. Then the 3D box regression net (c) classifies the objects and estimates 3D box parameters.}
	\label{3DBoxNetFig}
\end{figure}

Following the above works, a 3D object detection network is proposed to estimate the 3D bounding boxes and the categories of Lidar points in proposals. This network is composed of three subnetworks, which realizes semantic segmentation of Lidar points, coordinate transformation, and estimation of 3D bounding boxes in turn, as shown in Fig.\ref{3DBoxNetFig}. In this section, we describe the detection network in detail.

\subsection{Network Architecture}
After motion detection, all moving points are extracted and clustered to form independent objects. The 3D object detection network aims to further analyze the moving point clouds to provide more detailed semantic information. Inspired by PointNet series works \cite{qi2017pointnet,qi2017pointnet++,qi2018frustum}, our network utilizes a cascade framework to process the unordered point clouds. We also adopt the feature reuse method to reduce the network size and accelerate the processing speed.

\textbf{Segmentation Subnetwork.} 
A 3D moving object proposal obtained from motion detection contains at most one moving object, and it usually contains most of the object points as well as some background points. The segmentation subnetwork tries to assign each point a confidence score indicating the likelihood of belonging to the moving object. This likelihood score serves as the foundation for the subsequent classification and 3D bounding box regression subnetworks.

The subnetwork is an improved version over Frustum-PointNet \cite{qi2018frustum}, as shown in Fig.\ref{3DBoxNetFig}(a). Since the Lidar point cloud structure of each proposal is simple, it does not need a complex network to process. Therefore, it compresses the network scale by reducing the number of network layers and narrowing the network architecture. To maximize the utilization of features, it directly merges the features of each layer to fuse the global features and local features. In contrast to Frustum-PointNet, this segmentation subnetwork also does not introduce prior features of object categories, so it reduces the influence of prior errors to a certain extent. As a result, the subnetwork contains approximately 1.17M parameters, which is only one-third of the original Frustum-PointNet model. 

\begin{figure}[!t]
	\centering
	\subfloat[Center Calculation Diagram]{\includegraphics[scale=0.5]{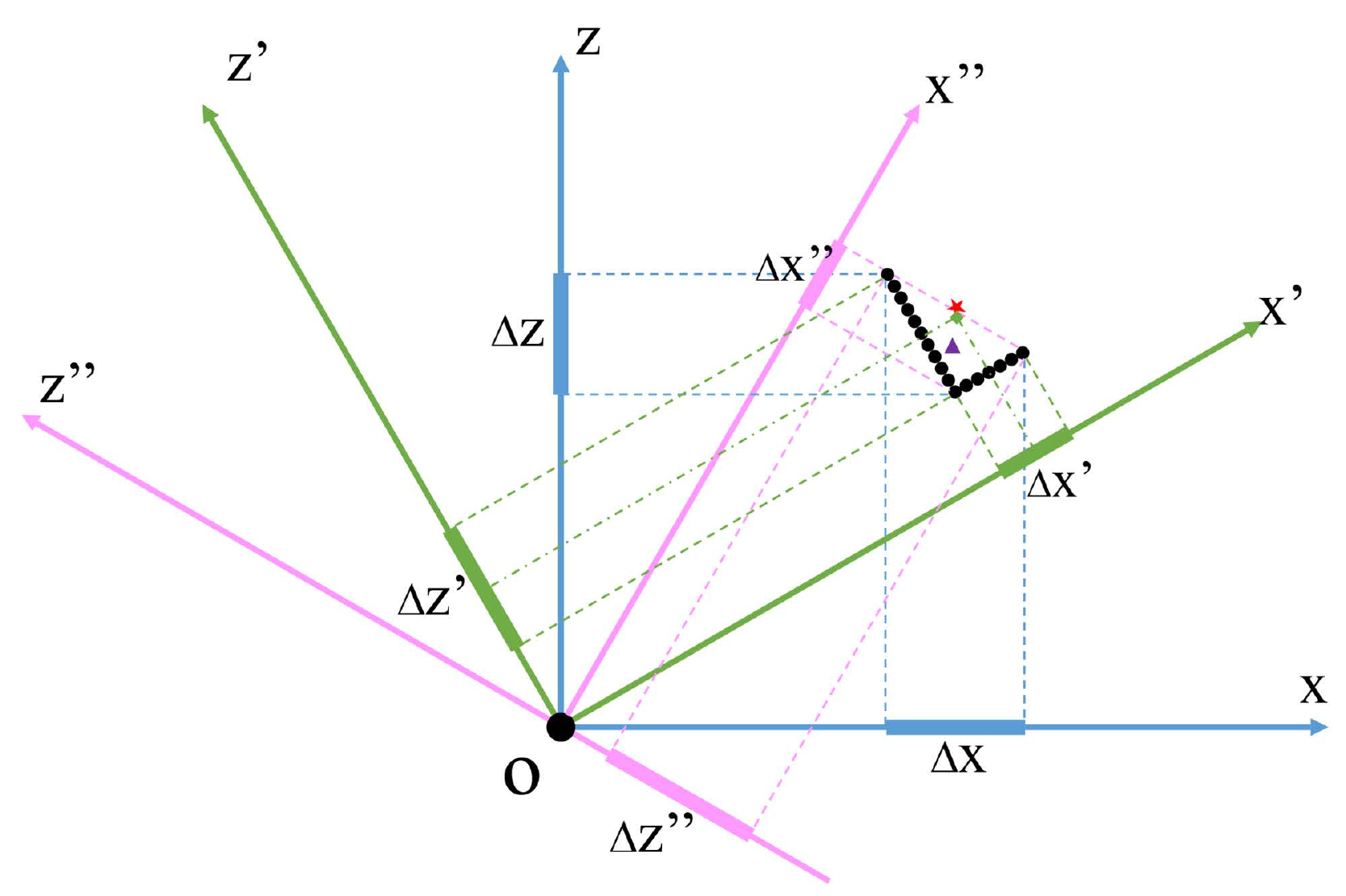}}\\
	\subfloat[Lidar Coordinate]{\includegraphics[scale=0.56]{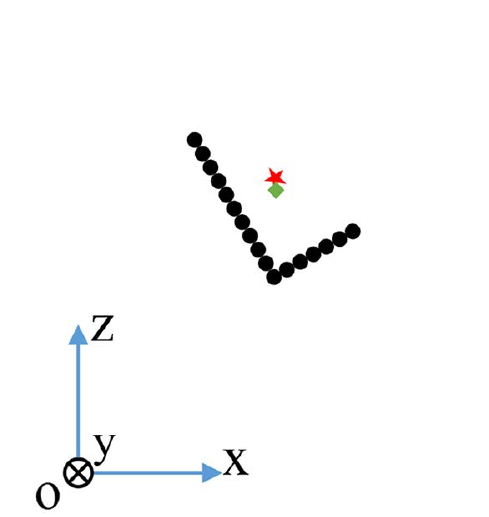}}\hspace{1.2pt}
	\subfloat[Points Coordinate]{\includegraphics[scale=0.56]{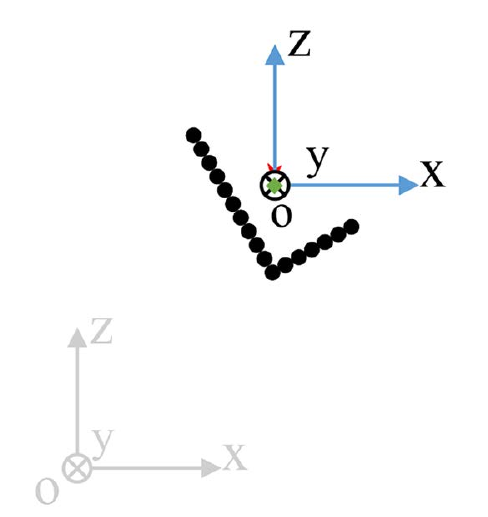}}\hspace{1.2pt}
	\subfloat[Object Coordinate]{\includegraphics[scale=0.56]{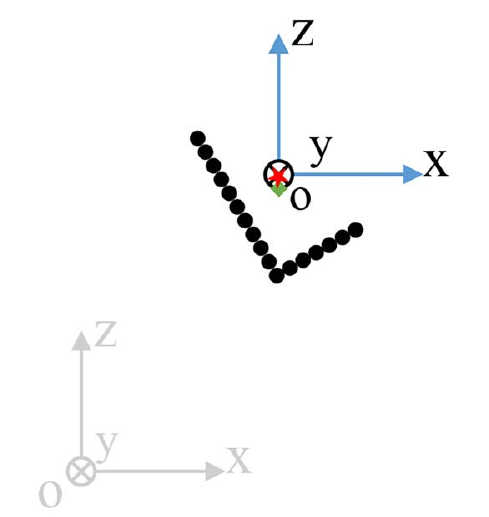}}\hspace{1.2pt}
	\caption{\textbf{Box center calculation and coordinate transformation.} (a) describes the computing process of box center, in which the purple triangle represents the average of all points; the green diamond represents our calculation results; the red pentagon represents the true center of an object. (b)$\sim$(c) illustrates the coordinate system transformation process from global to local, and (c)$\sim$(d) expresses the transformation effect of T-net. We can see the coordinate system approaches the truth object center step by step.}
	\label{CoordinateTransform}
\end{figure}

\textbf{T-Net.} T-Net is an intermediate unit of the 3D object detection network. It is a light-weight regression network that plays two roles: on the one hand, it transforms the foreground point clouds from the Lidar coordinate system to the object coordinate system (Fig.\ref{CoordinateTransform} (b)$\sim$(d)); on the other hand, it estimates the residuals between the point cloud center and the real object center. This process facilitates the subsequent 3D box estimation subnetwork calculating the box parameters.

\textbf{3D Box Estimation Subnetwork.}
Given the masked moving object points, this module predicts a moving object's state and category using an estimation network shown in Fig.\ref{3DBoxNetFig}(c). The classification subnetwork is a simple multi-layer perceptron based on PointNet. In contrast, it reduces the number of layers and merges local features and global features to achieve a good classification performance.

For the regression part, a moving object's state is described as $\bm{s} = \{c_x, c_y, c_z, h, w, l, \theta \}$, where $c_x, c_y, c_z$ are the object central coordinates; $h, w, l$ are the object sizes; $\theta$ is the object orientation. The center of 3D box is calculated according to Eq.\eqref{center}, which adopts a successive approximation strategy to improve the result accuracy. The initial object center $C_{seg}$ is roughly calculated based on the foreground points, and then the center residuals $\Delta C_{T\_net}$ and $\Delta C_{reg\_net}$ are calculated by T-Net and 3D Box Estimation Net to correct the final object center. Averaging the coordinates of all masked points is a naive approach for calculating the initial object center $C_{seg}$ \cite{qi2018frustum}. However, this result may deviate from the true object center owing to the unbalanced distribution of object points. Considering the distribution characteristics of object points, we develop an improved approach to reduce the impact of point clouds density on the initial object center, as shown in Fig.\ref{CoordinateTransform}. The x-z coordinate system rotates 30\degree and 60\degree counterclockwise around y-axis to form two new coordinate systems. 
Then the square sums of projection regions are calculated after object points are respectively projected to the x-axis and z-axis. The coordinate system with the smallest square sum is the optimal result. The median coordinates of the projection ranges, $c^{\alpha}_x$ and $c^{\alpha}_z$, are then transformed to the global coordinate system, written as:

\begin{equation}
\begin{aligned}
\left[ \begin{matrix}
	c^{'}_x\\
	c^{'}_z\\
\end{matrix} \right] = 
\left[ \begin{matrix}
	cos \alpha&		-sin \alpha\\
	sin \alpha&		cos \alpha\\
\end{matrix} \right] 
\left[ \begin{matrix}
	c^{\alpha}_x\\
	c^{\alpha}_z\\
\end{matrix} \right]   ,
\end{aligned}
\label{coordinateTrans}	
\end{equation}
where $\alpha \in \{0, \pi/6, \pi/3 \}$ is the rotation angle. Besides, $c^{'}_y$ is the average height of all masked points. So the initial object center $C_{seg}$ is written as $\left(c^{'}_x, c^{'}_x\right)$. The calculation process is described in Fig.\ref{CoordinateTransform}(a).

In order to improve the rotation and translation invariance of the approach, the global coordinate system is converted to the local coordinate system centered on $C_{seg}$. Although $C_{seg}$ is more accurate than the centroid of points, it is still far from the real center. Therefore, T-net is used to estimate the real center and calculate the center residuals $\Delta C_{T\_net}$. The box regression network then calculates the final center residuals $\Delta C_{reg\_net}$. The estimation object center $C_{est}$ is written as:

\begin{equation}
\begin{aligned}
C_{est} = C_{seg} + \Delta C_{T\_net} + \Delta C_{reg\_net} \, .
\end{aligned}
\label{center}	
\end{equation}

To predict the object size and angle, we adopt a common scheme that combined the classification result with the regression result. Note that the prior size of each object class is defined. The object category and the size residual, defined as $\left[s_{cls}\left( i \right), \delta h \left( i \right), \delta w \left( i \right),\delta l \left( i \right) \right]_{i=1}^{N_{cls}}$, are then jointly estimated. Similarly, we equally divide the angle, ranging from 0 to $\pi$,  into $N_{head}$ bins, and the network predicts the angle intervals and angle residuals, which can be written as $\left[h_{cls}\left( j \right), \delta \theta \left( j \right) \right]_{j=1}^{N_{head}}$.    

\subsection{Implementation Details}
This section describes the loss functions and some details in training the networks.

\subsubsection{Loss Function}
For this cascaded 3D object detection network, we design a multi-task loss function as follows: 

\begin{equation}
\begin{aligned}
\mathcal{L}_{total} = \delta \left(\bm{x}\right) \left( \alpha \mathcal{L}_{seg} + \gamma \mathcal{L}_{reg} \right) + \beta \mathcal{L}_{cls}  \, ,
\end{aligned}
\label{loss_total}	
\end{equation}
where $\alpha,\beta,\gamma$ are weight parameters used to adjust the contribution of different loss terms, and $\delta \left(\bm{x}\right)$ is an activation function. If the input proposal $\bm{x}$ is a positive sample, $\delta \left(\bm{x}\right) = 1$. It suggests that the segmentation and box regression loss functions are activated. On the other hand, if $\bm{x}$ belongs to the background, $\delta \left(\bm{x}\right) = 0$, and only the box classification loss is calculated. 

$\mathcal{L}_{seg}$ and $\mathcal{L}_{cls}$ represent the segmentation loss and the classification loss, which are modeled as cross-entropy losses. $\mathcal{L}_{reg}$ represents the 3D box regression loss, which is defined as:

\begin{equation}
\begin{aligned}
\mathcal{L}_{reg}\, =\, \mathcal{L}_{c\_Tnet} + \mathcal{L}_{c\_reg} + \mathcal{L}_{s\_cls}\,+\mathcal{L}_{s\_reg} + \mathcal{L}_{h\_cls} + \mathcal{L}_{h\_reg}  \, ,
\end{aligned}
\label{loss_reg}	
\end{equation}
where $\mathcal{L}_{c\_Tnet}$ and $\mathcal{L}_{c\_reg}$ are defined as smooth L1 loss, corresponding to the box center residuals $\Delta C_{T\_net}$ and $\Delta C_{reg\_net}$ in Eq.\eqref{center}. $\mathcal{L}_{s\_cls}$ and $\mathcal{L}_{h\_cls}$ are represented as cross-entropy functions that reflect the classification losses of size categories and the angle intervals. $\mathcal{L}_{s\_reg}$ and $\mathcal{L}_{h\_reg}$ are calculated using the Huber loss function, which respectively presents the regression losses of size and angle.

\subsubsection{Network Training}
We use the KITTI 3D object detection dataset \cite{Geiger2012CVPR} to train our network. To better simulate the output of moving object detection module, Lidar points are properly expanded to form a proposal according to the ground-truth. The positive data are augmented by rotation, translation, down-sampling, etc. In contrast, negative samples are generated by random sliding with different templates in the Lidar point cloud. The proportion of positive and negative samples is 1:1.

For this cascaded network, we adopt a stagewise training strategy. Firstly, we only use the augmented positive samples to train the segmentation subnetwork, and then fix the learned weight parameters of this module. Since the 3D box regression subnetwork and classification subnetwork share a part of the backbone network, we first pre-train the network using positive samples to obtain the initial parameters. Then both positive and negative samples are utilized to fine-tune these networks, in which only the positive data can be used to optimize the 3D box regression network by activation function $\delta \left(\bm{x}\right)$, while all kinds of data are applied for optimizing the classification network. Therefore, these networks can simultaneously estimate the 3D boxes and object categories.

\section{Experimental Evaluation}\label{sec:Experiments}
Following the above cascaded two modules, we perform several experiments on the modified KITTI benchmark \cite{Geiger2012CVPR} to verify the effectiveness of our approach.

\subsection{Dataset}
For moving object detection of traffic scenes, there is no specific dataset, including the widely-used KITTI benchmark. However, the KITTI benchmark includes datasets of similar tasks and provides images, Lidar point clouds, as well as the IMU data, which can be used to evaluate our approach after some modifications. We implement our experiments using a combination of the KITTI object detection dataset, the tracking dataset, and the raw dataset. The 3D object detection dataset consists of more than 7,000 training data with approximately 40,000 labeled objects, which mainly concentrate on three categories: \textit{car}, \textit{pedestrian} and \textit{cyclist}. As these data contain no ordinal information, they can just be used for training the 3D object detection network. Nevertheless, the tracking dataset and the raw dataset consist of sequential data, which can be utilized for motion detection. Based on the official groundtruth, we establish a motion detection dataset containing more than 1700 moving objects.

We use two types of metrics to evaluate the algorithm performance in 2D (Field of View, FoV), 3D and BEV (bird's eye view). Referring to the KITTI object dataset, average precision (AP) is used to evaluate the 3D detection performance. Following the official standard, the Intersection-over-Union (IoU) threshold is set to 0.7 for cars, and 0.5 for pedestrians and cyclists. To evaluate the motion detection performance, we use three indicators: precision (PRE), recall (REC) and F1-score. Among them, PRE indicates the proportion of the detected moving objects with respect to all the detected targets; REC represents the ratio of the detected moving targets versus all the moving targets; F1-score is an integrated indicator that can be regarded as a weighted average of precision and recall.

\subsection{Parameter Settings}

\begin{figure}[t]
	\centering
	\subfloat[Precision --- Weights]{\includegraphics[scale=0.18]{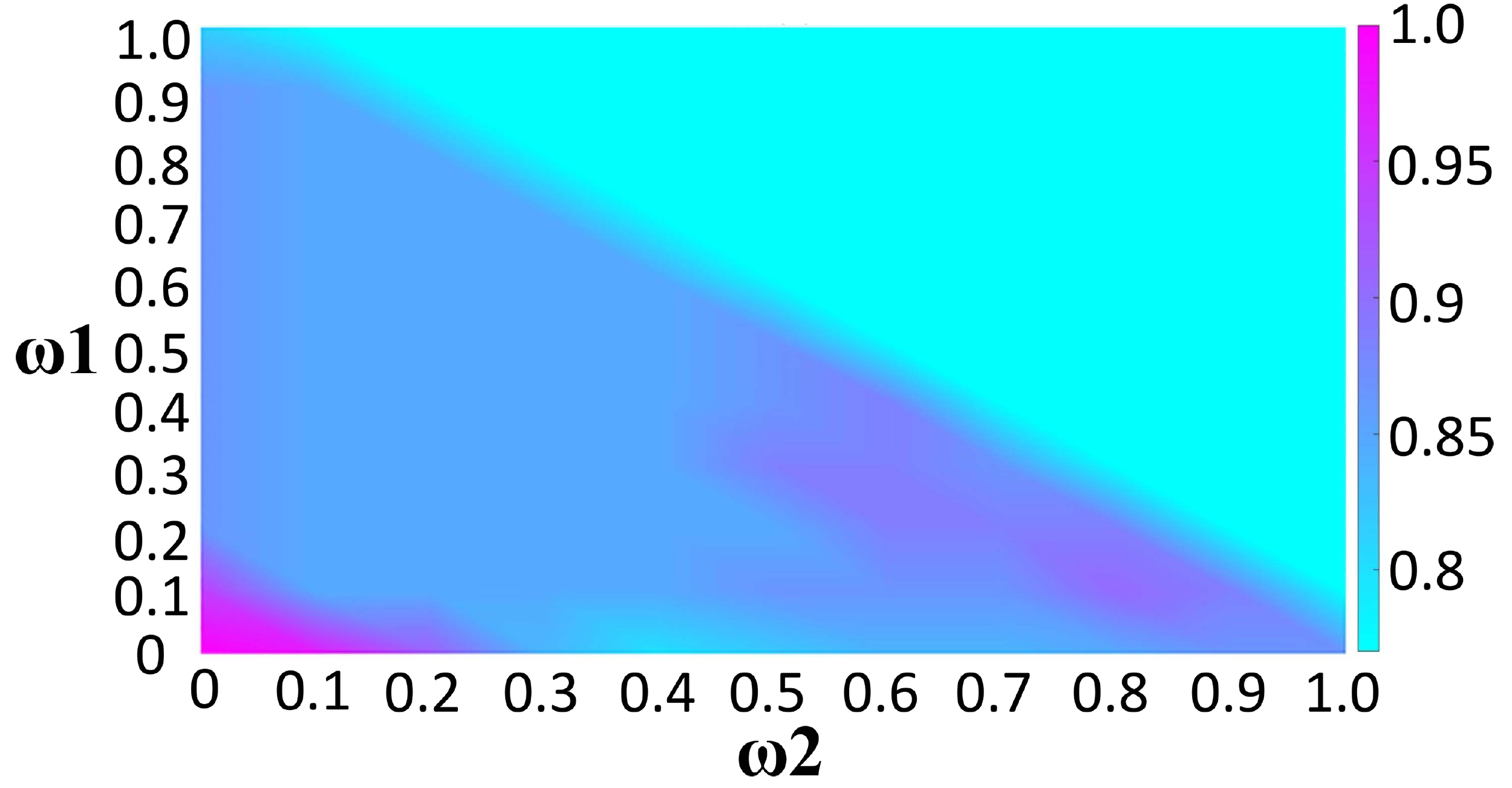}}
	\subfloat[Recall --- Weights]{\includegraphics[scale=0.176]{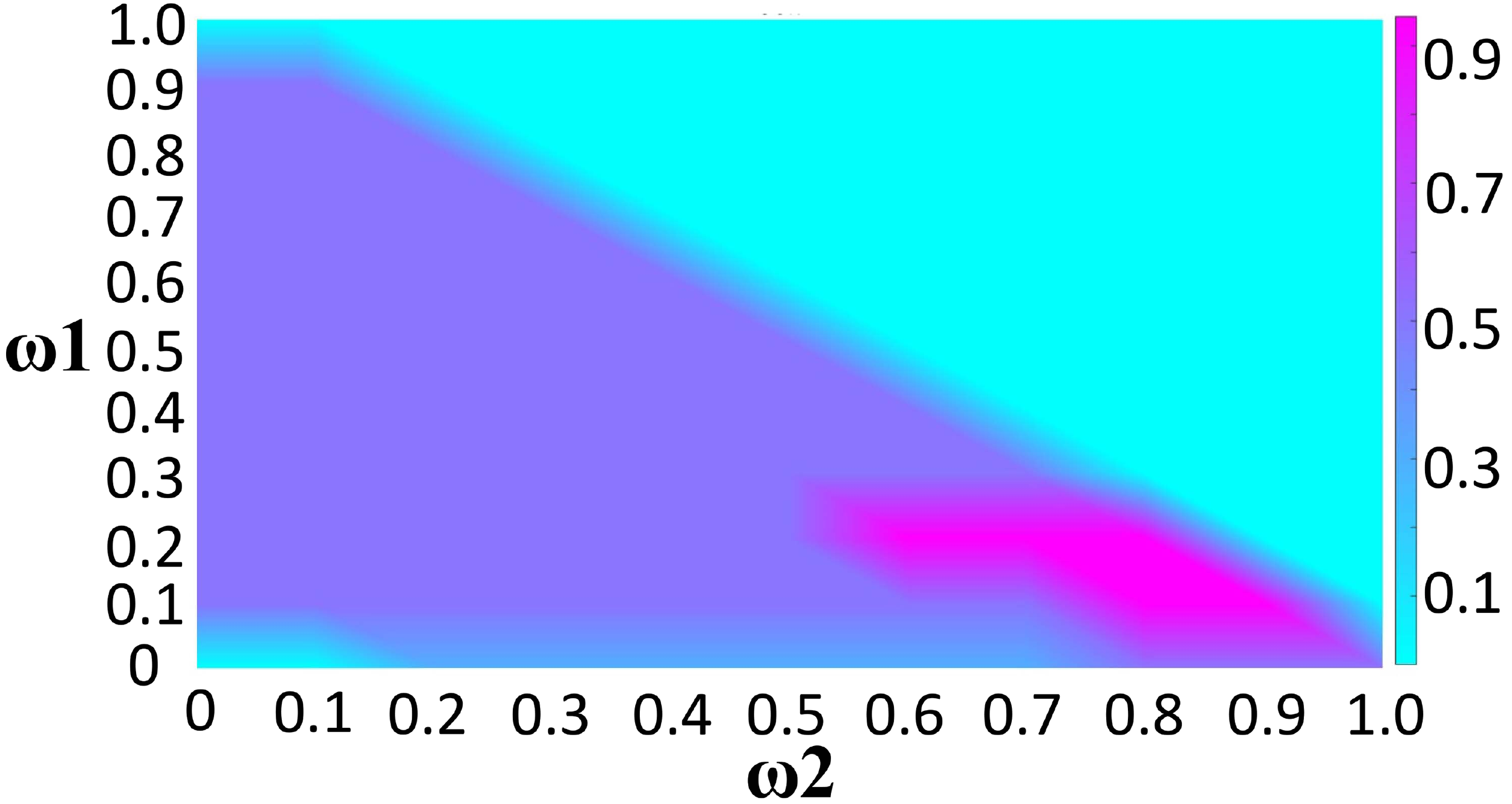}}  \\
	\subfloat[F1 Score --- Weights]{\includegraphics[scale=0.32]{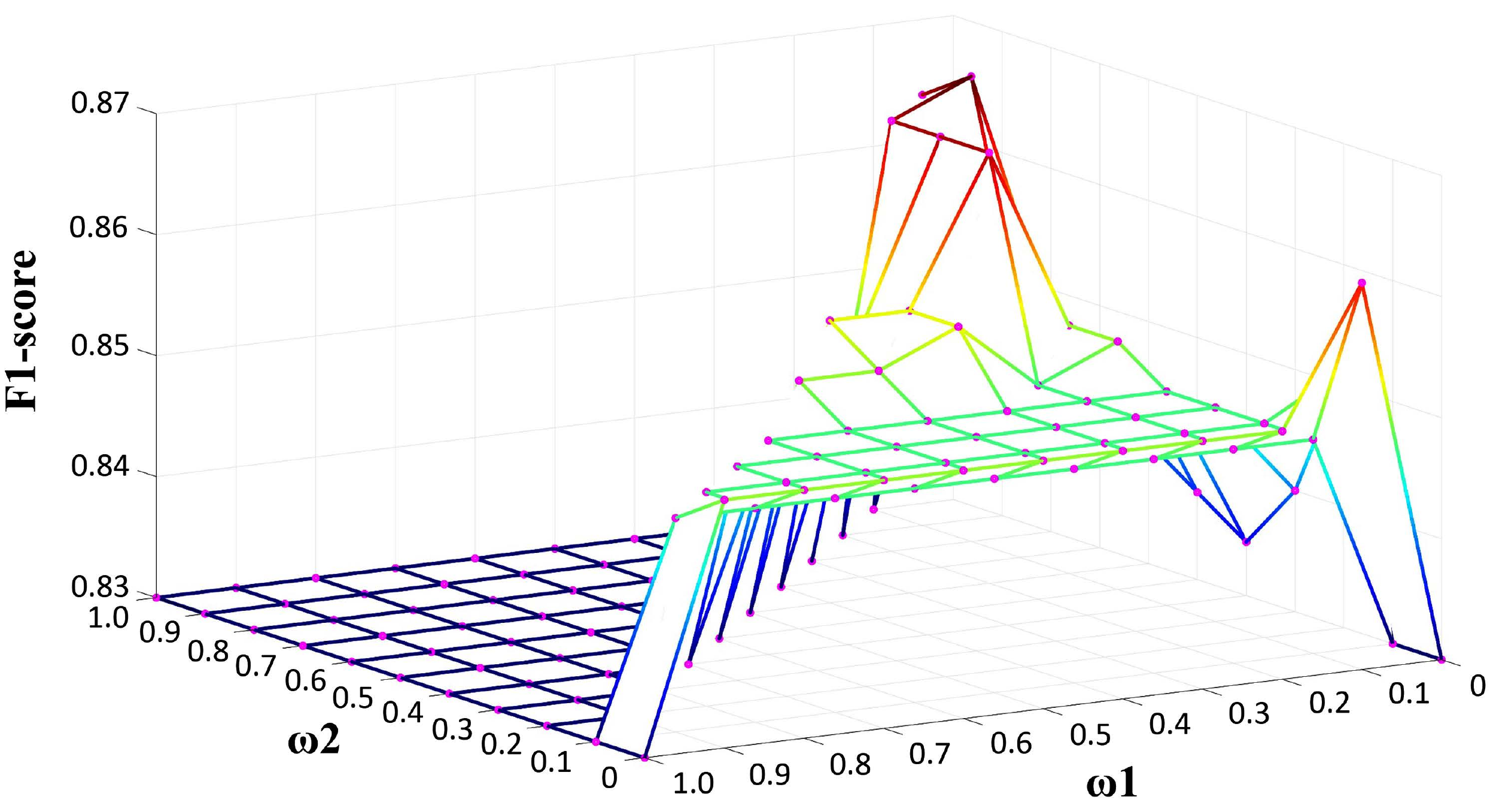}} \\
	\caption{\textbf{Evaluation results with different weights.} (a) shows the precision scores under different weight combinations; (b) shows the recall score at different weights; (c) illustrates the relation between F1-scores and combined weights.}
	\label{ParaEvl}
\end{figure} 

In the EMD-based motion detection model, three weight parameters balance the matching scores of different features. These parameters directly affect the detection performance of dynamic targets. With the constraint $\bm{\omega_{1}}+\bm{\omega_{2}}+\bm{\omega_{3}}=\bm{1}$ in Eq.\eqref{Energy}, we perform a series of experiments on 5 raw data sequences to select the optimal parameters. 

Fig.\ref{ParaEvl} shows the evaluation results with different weight groups. It can be observed from Fig.\ref{ParaEvl} (a) and (b) that the parameters achieving higher recall cannot obtain higher precision. However, the F1-scores shown in (c) can help us choose the optimum parameters. According to the maximum F1-score, $\{\bm{\omega_{1}},\bm{\omega_{2}},\bm{\omega_{3}}\}$ are set to $\{0.1,0.8,0.1\}$. With these parameters, the method could obtain a high recall rate while keeping the false alarm rate to a low level. In addition, the experiment shows that the shape of objects has good stability under the representation of Lidar data. At the same time, the height feature map and Gaussian feature map further improve the performance of the approach.

Through data analysis and qualitative experiments, other parameters of the EMD-based motion detection model are set as follows: the time constant $\tau$ in Eq.\eqref{first-order LPF} is set to $2$; the maximum connection distance $R$ is set to $10$; the patch size $m$ is $21$; $\left\lbrace l, p, q, M, N \right\rbrace = \left\lbrace 15, 0.56, -0.01, 200, 350 \right\rbrace$ in Eq.\eqref{velocity} and Eq.\eqref{Filter}.

\subsection{Performance Evaluation}
We test the proposed algorithm on motion detection dataset and analyze the results from the following aspects: 1) the structure of the EMD-based model; 2) the fusion of frames with different intervals; 3) the architecture of the 3D detection network; 4) the comparison with other state-of-the-art motion detection approaches.

\textbf{The EMD-based model.} Based on the EMD principle discovered by neurobiologists, we are the first to extend and apply it to 3D Lidar point cloud data. In biological research, the motion scores are usually calculated based on the binary information of pulse signal or the gray information of images, which is similar to just using $E^{'}_{1}$ in Eq.\eqref{Energy}. We treat this approach as the baseline approach and name it Coarse\_EMD method. By comparison, the approach proposed in Sec.\ref{sec:EMD_main} not only calculates the motion direction in a coarse to fine fashion, but also integrates more 3D information. We named this approach as C2F\_EMD. Due to the incorporation of more 3D information, C2F\_EMD should be more effective than Coarse\_EMD in theory.

\begin{table*}[!t]
	\centering
	\caption{Comparison experiment results of different EMD structures and different frames fusion.}
	\label{Tab_1}
	\renewcommand\arraystretch{1.5}
	\resizebox{\textwidth}{!}{
	\begin{tabular}{c ||
	                     p{1.2cm}<{\centering} | p{1.2cm}<{\centering} | p{1.4cm}<{\centering} || 
						 p{1.2cm}<{\centering} | p{1.2cm}<{\centering} | p{1.4cm}<{\centering} || 
						 p{1.2cm}<{\centering} | p{1.2cm}<{\centering} | p{1.4cm}<{\centering} }
		\Bhline
		\multirow{2}{*}{Method}&  
		\multicolumn{3}{c||}{3D}& \multicolumn{3}{c||}{BEV}& \multicolumn{3}{c}{2D}\cr
		\cline{2-10}
		&PRE &\bf{REC} &F1-score &PRE &\bf{REC} &F1-score &PRE &\bf{REC} &F1-score\cr
		\hline
		Coarse\_EMD\_1 &{0.6128}     &{0.7336}     &{0.6678}    
				  &{0.6646} 	&{0.7956} 	  &{0.7242} 	  
				  &{0.6921}     &{0.8225}     &{0.7517} \\  
		Coarse\_EMD\_2 &{0.5281}     &{0.8577}     &{0.6537}    
			      &{0.5551} 	&{0.8915} 	  &{0.6842} 	  
			      &{0.5692}     &{0.9239}     &{0.7044} \\
		Coarse\_EMD\_3 &{0.5210}     &\textcolor[rgb]{1,0,0}{\bf{0.8613}} &{0.6493}    
				  &{0.5364}  	&{0.8869}    &{0.6685} 	  
				  &{0.5614}     &{0.9250}    &{0.6987}  \\
		Coarse\_EMD\_4 &{0.5156}     &{0.8431} 	 &{0.6399}    
		          &{0.5424}  	&{0.8869} 	 &{0.6731} 	  
		          &{0.5667}     &{0.9239}    &{0.7025}  \\
		\hline
		C2F\_EMD\_1 &{0.7551}    &{0.8102} 	 &{0.7817}    
		       &{0.7959}    &{0.8540} 	 &{0.8239}    
		       &{0.8333}    &{0.8877}    &{0.8596}        \\  
		C2F\_EMD\_2 &\textcolor[rgb]{0,0.78,0.55}{\bf{0.7752}} &{0.8431} 	 &\textcolor[rgb]{0,0,1}{\bf{0.8077}}
		       &\textcolor[rgb]{0,0.78,0.55}{\bf{0.8188}} &{0.8905} 	 &\textcolor[rgb]{0,0,1}{\bf{0.8531}} 
		       &\textcolor[rgb]{0,0.78,0.55}{\bf{0.8523}} &{0.9203}    &\textcolor[rgb]{0,0,1}{\bf{0.8850}} \\
		C2F\_EMD\_3 &{0.7428}    &{0.8431} 	 &{0.7898}    
		       &{0.7814}  	&{0.8869} 	 &{0.8308}    
		       &{0.8205}    &\textcolor[rgb]{1,0,0}{\bf{0.9275}} &{0.8707} \\
		C2F\_EMD\_4 &{0.7196}    &{0.8431} 	 &{0.7765}    
		       &{0.7632}  	&\textcolor[rgb]{1,0,0}{\bf{0.8942}} &{0.8235}
		       &{0.7888}    &{0.9203}    &{0.8495} \\
		\Bhline
	\end{tabular}}
\end{table*}

Table \ref{Tab_1} shows the comprehensive evaluation results of all object categories from the perspective of 2D, 3D and BEV. Although the Coarse\_EMD is comparative to C2F\_EMD in the recall evaluation index, the C2F\_EMD method is much better than Coarse\_EMD in terms of precision. This illustrates that the Coarse\_EMD method obtains a high recall of moving objects by increasing the number of detected objects. In contrast, the C2F\_EMD method keeps both the false detection rate and the miss detection rate to a low level. Overall, the C2F\_EMD approach performs better than Coarse\_EMD.

\begin{figure*}[!t]
	\centering
	\subfloat[Total]{\includegraphics[scale=0.24]{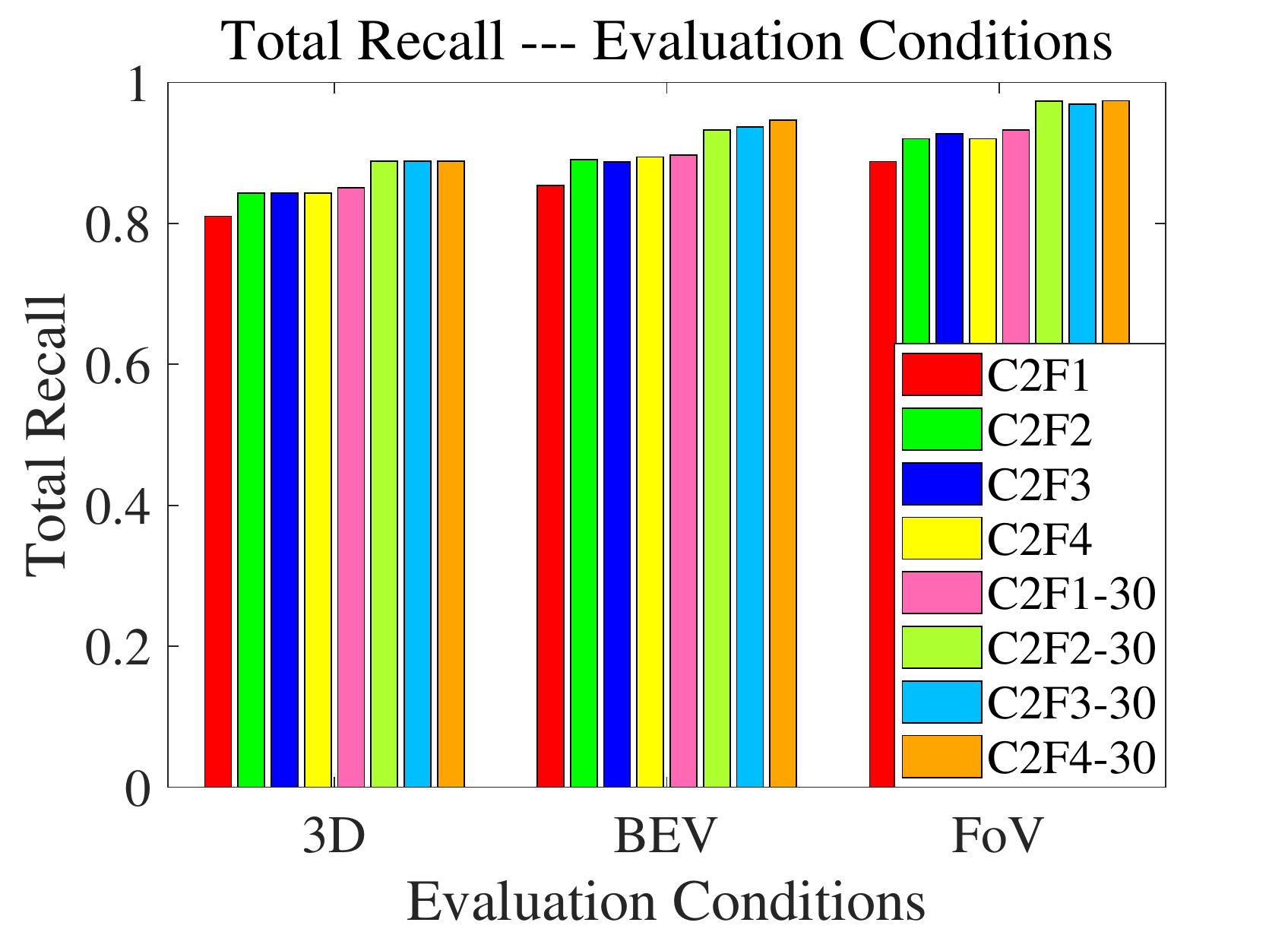}}
	\subfloat[Car]{\includegraphics[scale=0.24]{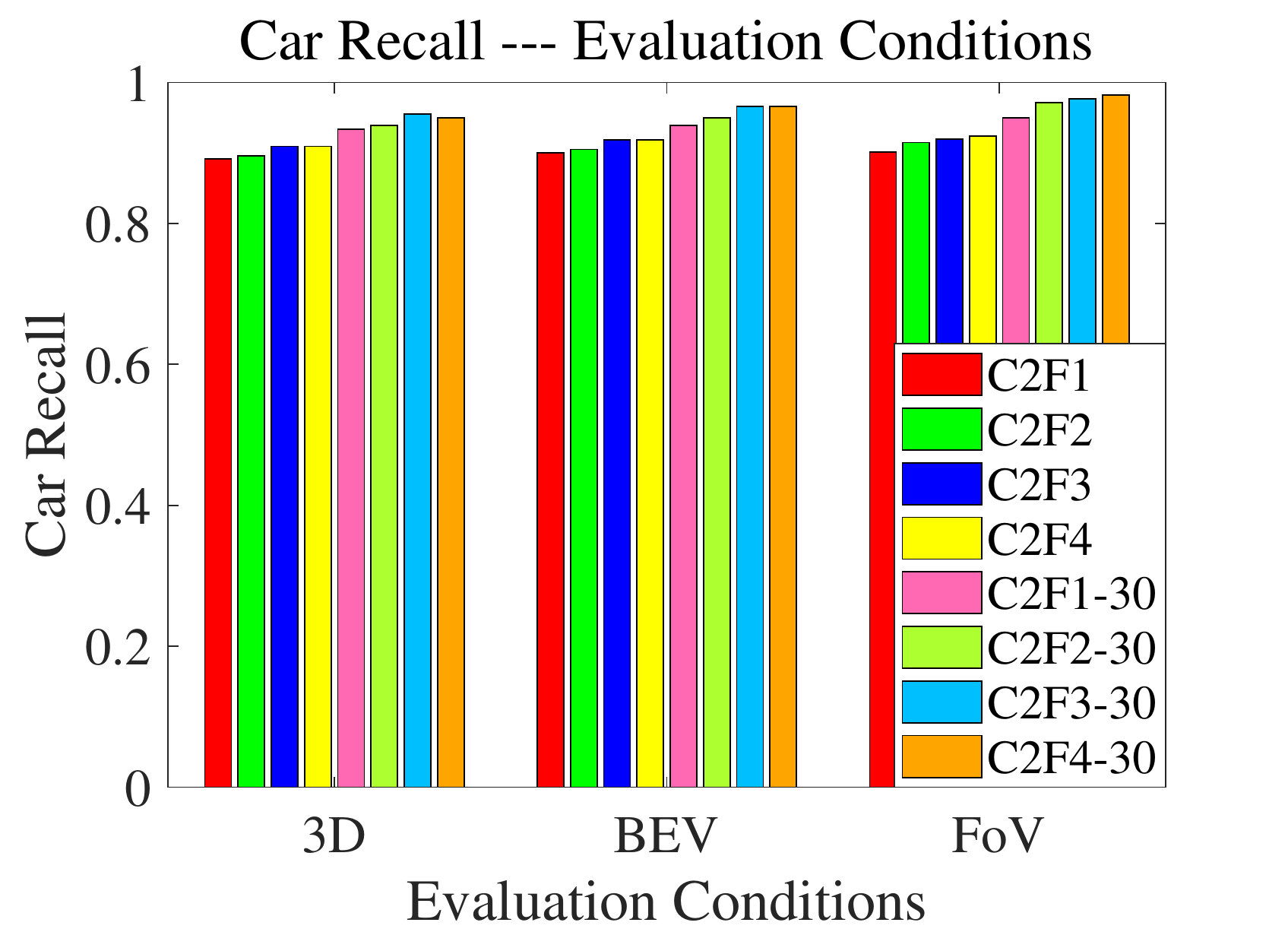}}
	\subfloat[Pedestrian]{\includegraphics[scale=0.24]{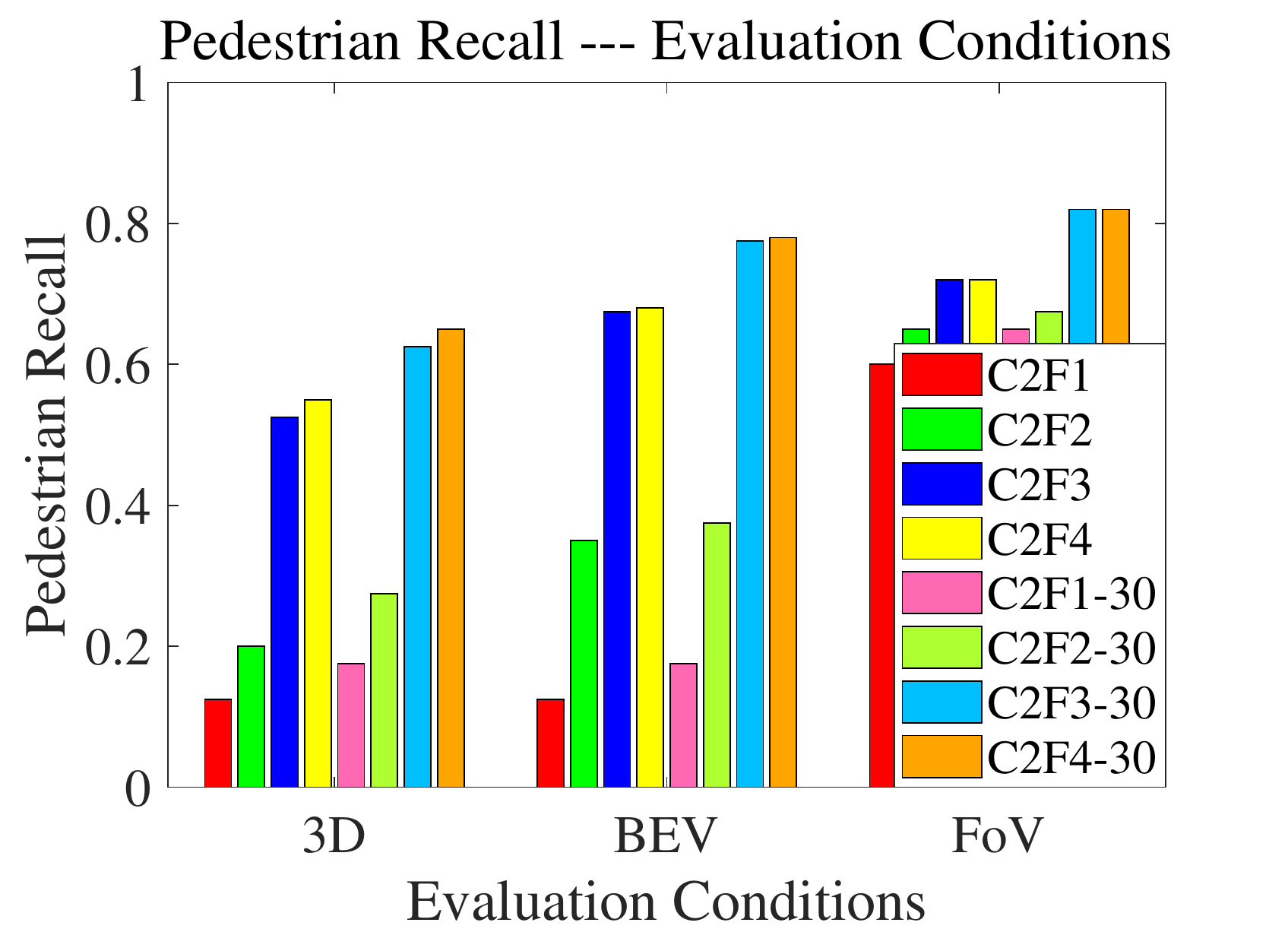}}
	\subfloat[Cyclist]{\includegraphics[scale=0.24]{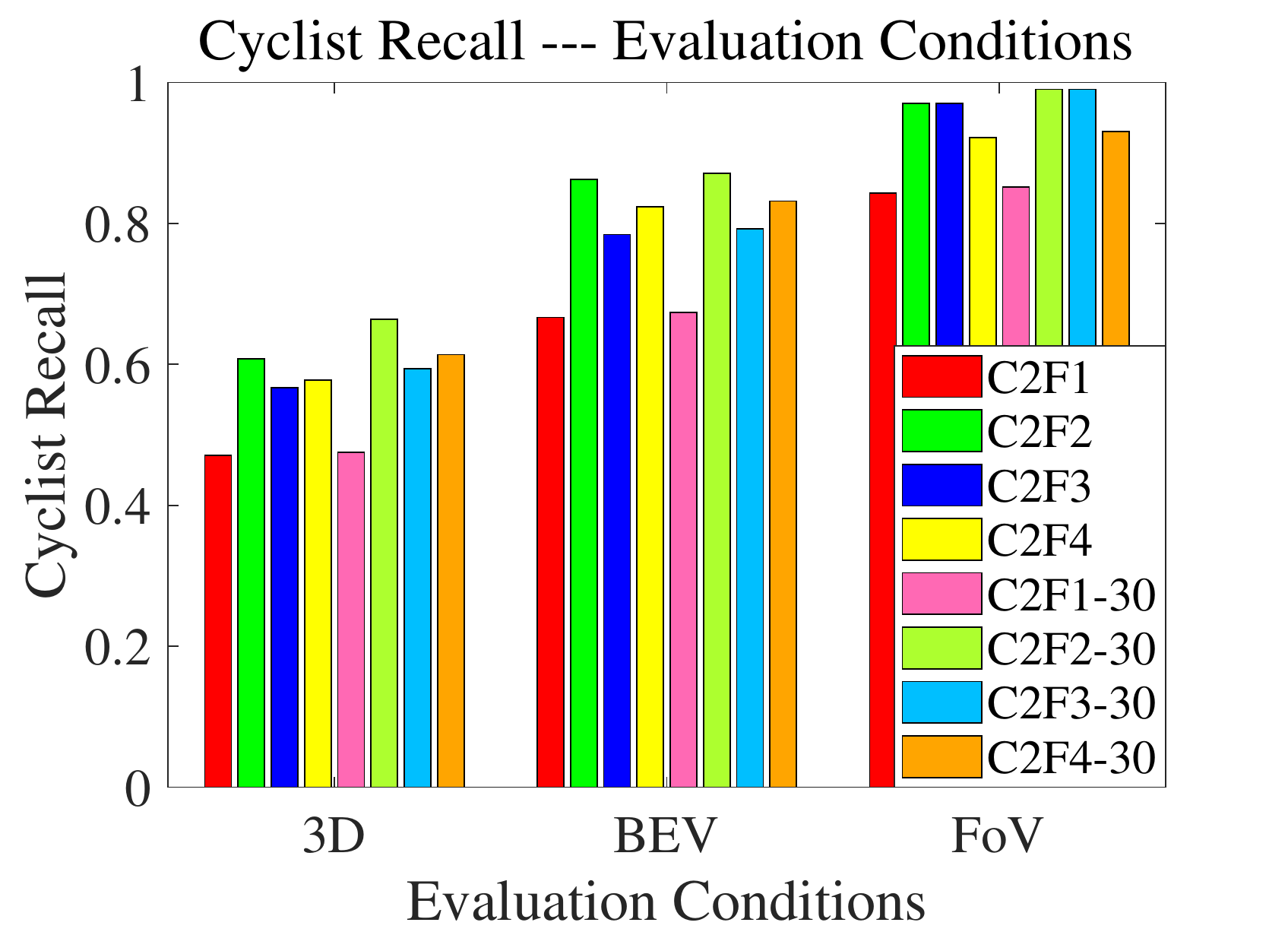}}
	\caption{\textbf{The recall of motion objects under different distance range.} (a) describes the total recall of all object categories. (b)$\sim$(d) respectively describes the recall of car, pedestrian and cyclist.}
	\label{recall_distance}
\end{figure*}

\textbf{Multi-frame Fusion.} Since the speed of moving objects varies greatly in traffic scenes, we try to fuse different numbers of consecutive frames to detect the dynamic traffic participants. Four multi-frame fusion methods (utilizing 2, 3, 4 or 5 consecutive frames) are evaluated, and the results are denoted as  *\_1, *\_2, *\_3 and *\_4 in Table \ref{Tab_1} and Fig. \ref{recall_distance}, respectively.

From the table, it can be seen that the recall rate increases as the number of fusion frames increases, but the precision declines. Through visualizing the detection results, it is found that the increase of recall is attributed to the improvement in detecting more low-velocity objects. A low-velocity object moves farther in multiple frames than that in two sequential frames, which makes it easier to detect them. However, the computational time also increases with the number of fused frames. To balance efficiency and performance, we finally use 3 consecutive frames for fusion.

\begin{table*}[t]
	\centering
	\caption{Comparison experiment results of different detection networks with various frames fusion in 3D view (IoU = 0.7).}
	\label{Tab_2}
	\renewcommand\arraystretch{1.5}	
	\resizebox{\textwidth}{!}{
	\begin{tabular}{c || c | c | c || c | c | c || c | c | c || c | c | c}
		\Bhline
		\multirow{2}{*}{Method}&  
		\multicolumn{3}{c||}{Total}& \multicolumn{3}{c||}{Car}& \multicolumn{3}{c||}{Pedestrian}& \multicolumn{3}{c}{Cyclist} \cr
		\cline{2-13}
		&PRE &\bf{REC} &F1 &PRE &\bf{REC} &F1 &PRE &\bf{REC} &F1 &PRE &\bf{REC} &F1\cr
		\hline
		F-PointNet*\_1 &{0.5125} &\bf{0.7667} &\bf{0.6143} &{0.6440} &{0.8610}    &{0.7369} &{0.1053} &{0.0750}    &{0.0876} &{0.4783} &\bf{0.4314} &\bf{0.4536} \\  
		Ours\_1       &{0.5112} &{0.7625}    &{0.6121} &{0.6693} &\bf{0.8984} &\bf{0.7671} &{0.1389} &\bf{0.1250} &\bf{0.1316} &{0.2917} &{0.2745}    &{0.2828} \\
		\hline
		F-PointNet*\_2 &{0.5527} &{0.7883}    &{0.6498} &{0.6459} &{0.8877}    &{0.7477} &{0.3000} &\bf{0.2500} &\bf{0.2727} &{0.5385} &\bf{0.5490} &\bf{0.5437} \\
		Ours\_2       &{0.5533} &\bf{0.7901} &\bf{0.6508} &{0.6880} &\bf{0.9198} &\bf{0.7872} &{0.2222} &{0.2000}    &{0.2105} &{0.3571} &{0.3922}    &{0.3738} \\
		\hline
		F-PointNet*\_3 &{0.5184} &\bf{0.8208} &\bf{0.6355} &{0.6117} &{0.8930} 	  &{0.7261} &{0.5056} &\bf{0.5825} &\bf{0.5413} &{0.5472} &\bf{0.5686} &\bf{0.5577} \\
		Ours\_3       &{0.5040} &{0.7917}    &{0.6159} &{0.6439} &\bf{0.9305} &\bf{0.7611} &{0.5056} &{0.5250}    &{0.5151} &{0.2679} &{0.2941}    &{0.2804} \\
		\hline
		F-PointNet*\_4 &{0.4606} &{0.8042} &{0.5857}    &{0.5680} &{0.8930}    &{0.6944} &{0.5157} &{0.5250}    &{0.5203} &{0.5254} &\bf{0.6078} &\bf{0.5636} \\
		Ours\_4       &{0.4738} &\bf{0.8292} &\bf{0.6030} &{0.5884} &\bf{0.9251} &\bf{0.7193} &{0.5333} &\bf{0.5500} &\bf{0.5415} &{0.4519} &{0.4725}    &{0.4620} \\
		\Bhline
	\end{tabular}}
\end{table*}

\begin{figure*}[!t]
	\centering
	\subfloat[2D]{\includegraphics[scale=0.28]{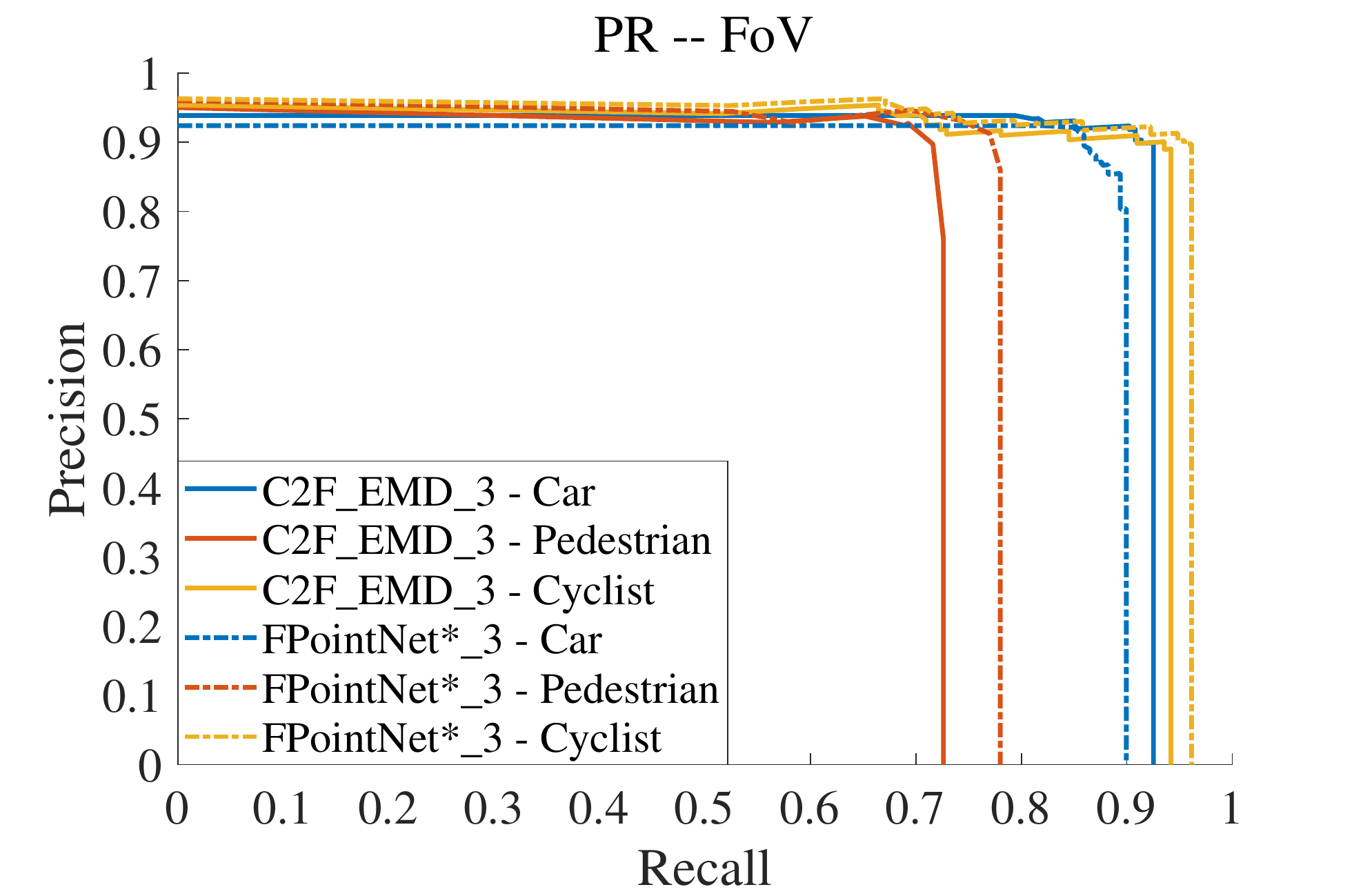}}
	\subfloat[BEV]{\includegraphics[scale=0.28]{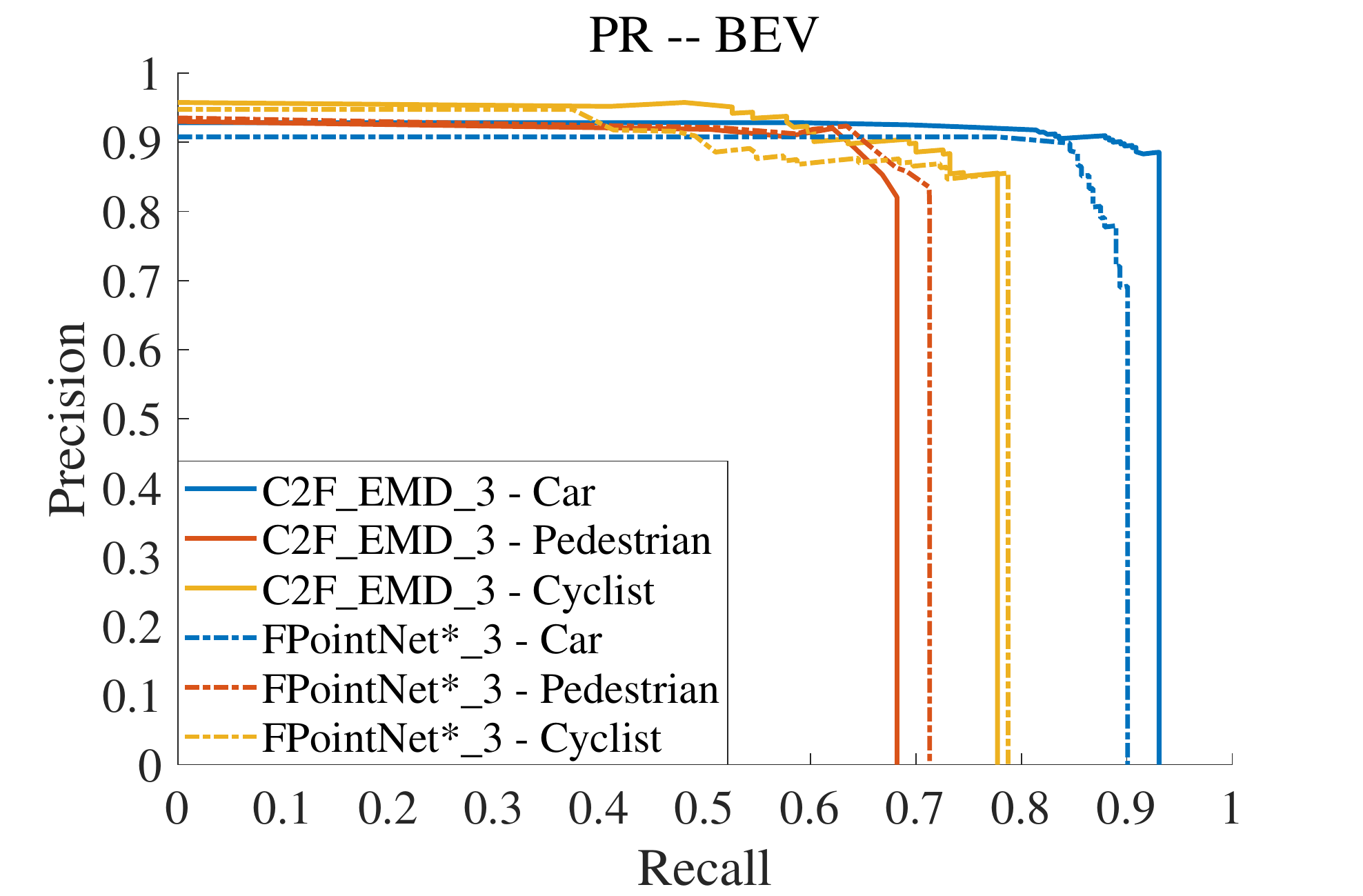}}
	\subfloat[3D]{\includegraphics[scale=0.28]{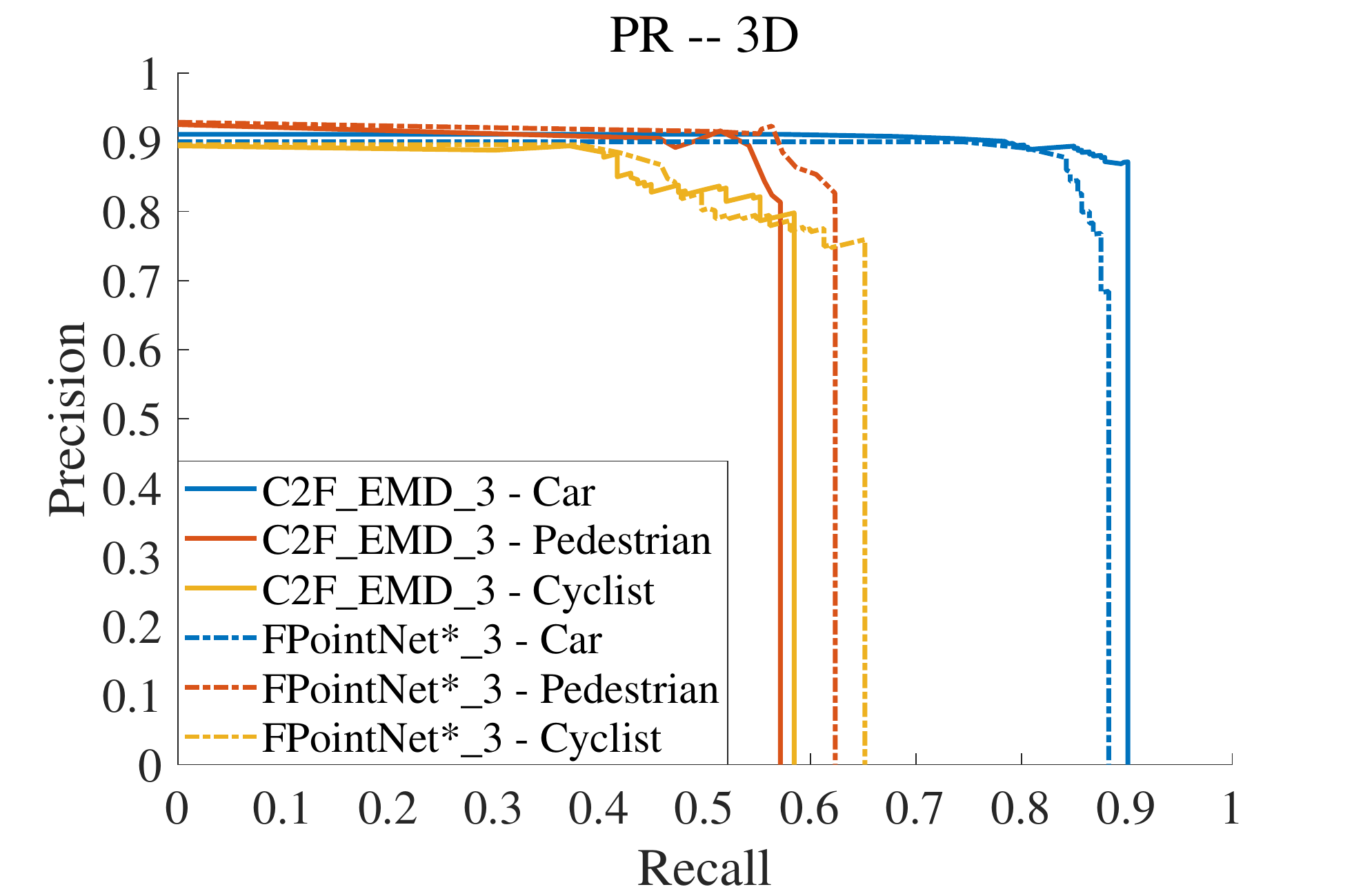}}
	\caption{\textbf{PR curves of different networks (IoU = 0.5).} (a)$\sim$(c) respectively represent the PR curves at the perspectives of 2D, BEV and 3D. In these graphs, the solid line indicates the result of the our method, and the dotted line indicates the result of F-PointNet.}
	\label{PR_Network}
\end{figure*} 

\textbf{Detection Network.} In order to verify the performance of our 3D detection network, a more rigorous standard, $IoU = 0.7$, is used for quantitative testing on the motion dataset. The baseline method is the Frustum-PointNet \cite{qi2018frustum}, which directly processes the point clouds and obtains a good performance in the 3D detection dataset. Under the same evaluation protocols, we conduct experiments and list the evaluation results in Table \ref{Tab_2}.

As shown in Table \ref{Tab_2}, our network is superior to F-PointNet in the car detection task. For cyclist detection, F-PointNet is more effective than our method. In terms of pedestrian detection, the performance of these two methods is similar. Fig.\ref{PR_Network} shows the PR curves of these two networks. It can be seen that our approach performs better than F-PointNet in car detection, but slightly inferior to F-PointNet in pedestrian and cyclist detection. The reason for this result is probably that there are few Lidar points to describe cyclists and pedestrians, which makes our simplified network's feature learning ability not as good as the complex F-PointNet. Besides, the PR curve of our approach is very flat, which proves the robustness of our network. Overall, with only one-third parameters of F-PointNet, our network performs on-par with F-PointNet, and shows a good  balance between performance and efficiency.

\textbf{Comprehensive Performance.} We compare the proposed algorithm with several state-of-the-art motion detection algorithms, including triTrack\cite{Lenz2011Sparse}, DSCNOD\cite{Xiao2017Dense},  DSCMOD*\cite{Xiao2018PhDthesis}, SOF\cite{sevilla2016optical}, DATMO\cite{asvadi2015detection} and lidarFMOD\cite{Xiao2018PhDthesis}. Among them, \textit{triTrack}, \textit{DSCNOD}, \textit{DSCMOD*} and \textit{SOF} detect moving objects based on optical flow and scene flow of images; \textit{DATMO} detects dynamic objects based on point clouds, and \textit{lidarFMOD} fuses Lidar point clouds and images to detect moving objects. These comparison methods are implemented in a PC with an Intel i7-7700HQ CPU and an NVIDIA GTX 1050Ti GPU.

The results are shown in Table \ref{Tab_3}. It can be seen that both the recall and precision of our algorithm are significantly better than the other comparative methods. Our approach improves the recall rate while maintaining a low false detection rate. It proves the effectiveness of the proposed algorithm. Besides, the average computational time of the proposed approach is around 500ms, which is much faster than most of the comparison approaches. It proves that our coarse-to-fine motion detection algorithm and the improved network efficiently reduce the computational cost. Although our approach does not run in real-time, it can be further accelerated by using parallel computing, which we considered as future work.

\begin{table}[!t]
	\centering
	\caption{Motion detection results compared with state-of-the-art methods in FoV.}
	\label{Tab_3}
	\renewcommand\arraystretch{1.5}
	\begin{tabular}{c || c | c | c | c}
		\Bhline
		Method &Precision &Recall &F1-Score &Run-time(s)\\
		\hline
		triTrack+viso2\cite{Lenz2011Sparse} &{0.4617}   &{0.4976}     &{0.4790} &\bfseries{0.451}\\  
		DSCMOD\cite{Xiao2017Dense}   &{0.1273}   &{0.4422}     &{0.1977} &{20.046}\\
		DSCMOD*\cite{Xiao2018PhDthesis}  &{0.1096}   &{0.5189}     &{0.1810} &{20.332}\\
		SOF\cite{sevilla2016optical} &{0.4405}   &{0.6851}     &{0.5362} &{72}\\
		DATMO(lidar)\cite{asvadi2015detection} &{0.1803}   &{0.6297}     &{0.2804} &{6.833}\\
		FMOD(lidar)\cite{Xiao2018PhDthesis} &{0.4401}   &{0.5719}     &{0.4974} &{16.447}\\
		\bfseries{Ours}     &\textcolor[rgb]{0,0.78,0.55}{\bfseries{0.8523}}   &\textcolor[rgb]{1,0,0}{\bf{0.9203}}   &\textcolor[rgb]{0,0,1}{\bfseries{0.8850}} &{0.502}\\
		\Bhline
	\end{tabular}
\end{table} 

\begin{figure*}[!t]
	\centering	
	\subfloat[]{\includegraphics[scale=0.3]{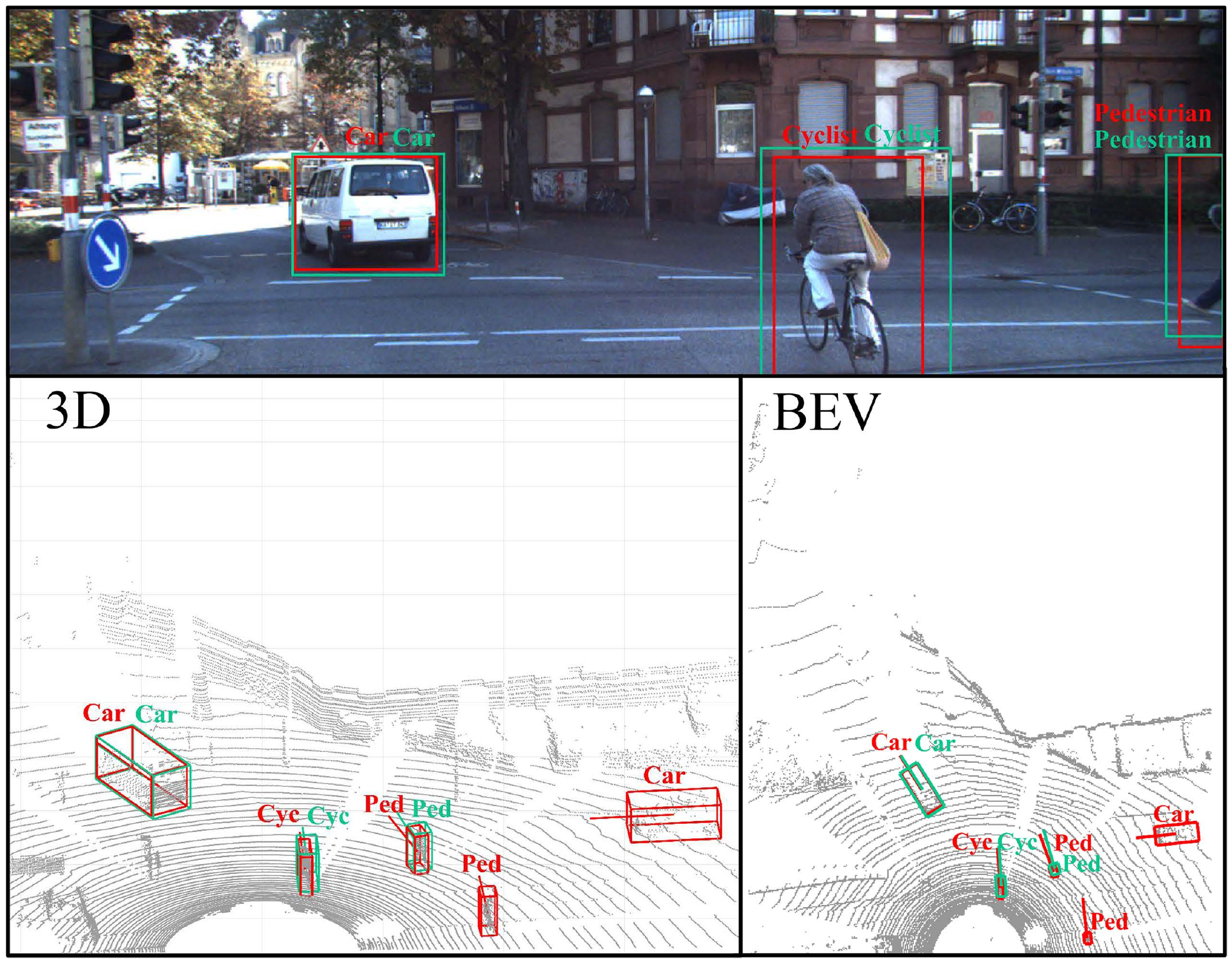}} \hspace{5pt}
	\subfloat[]{\includegraphics[scale=0.3]{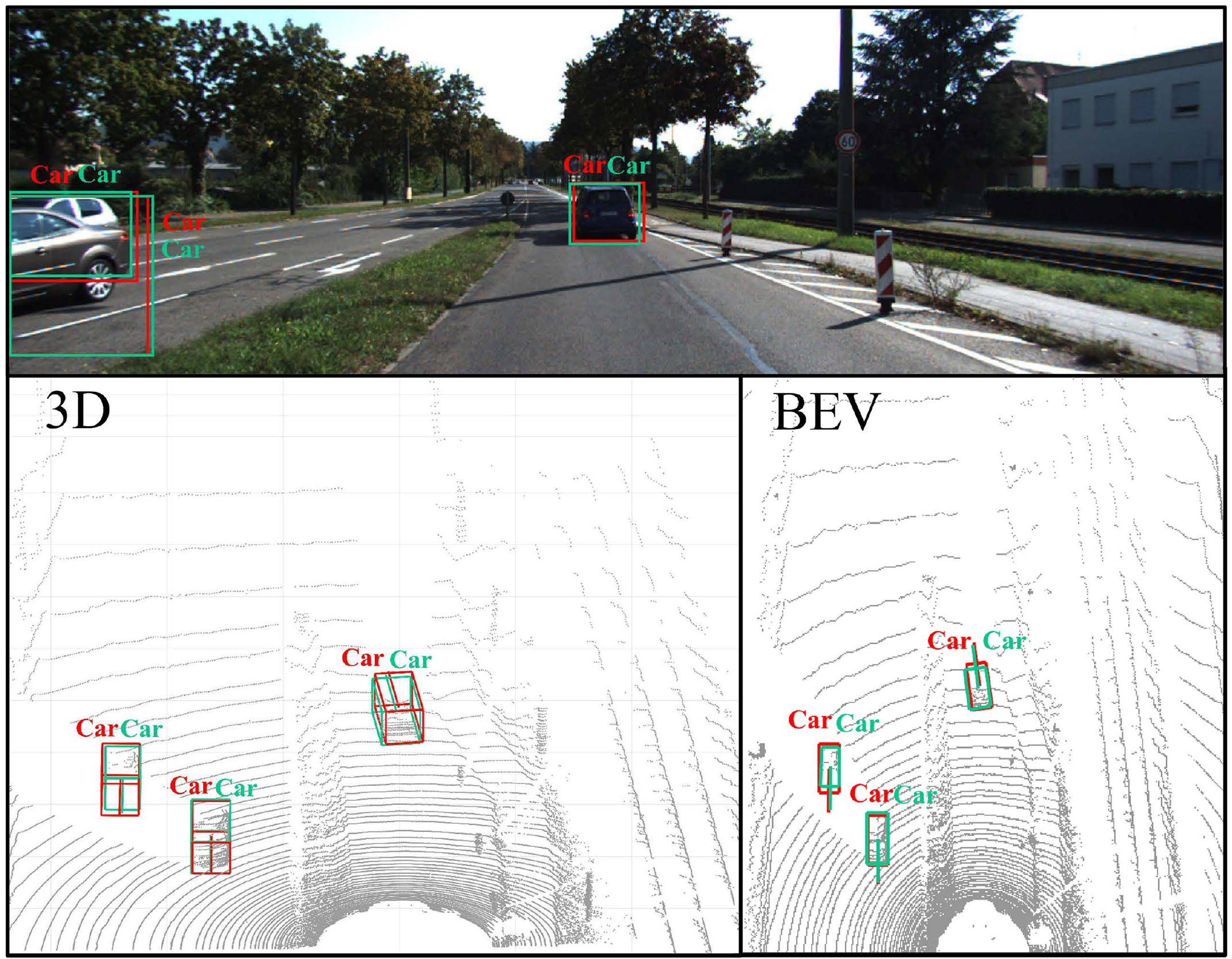}}\\
	\subfloat[]{\includegraphics[scale=0.3]{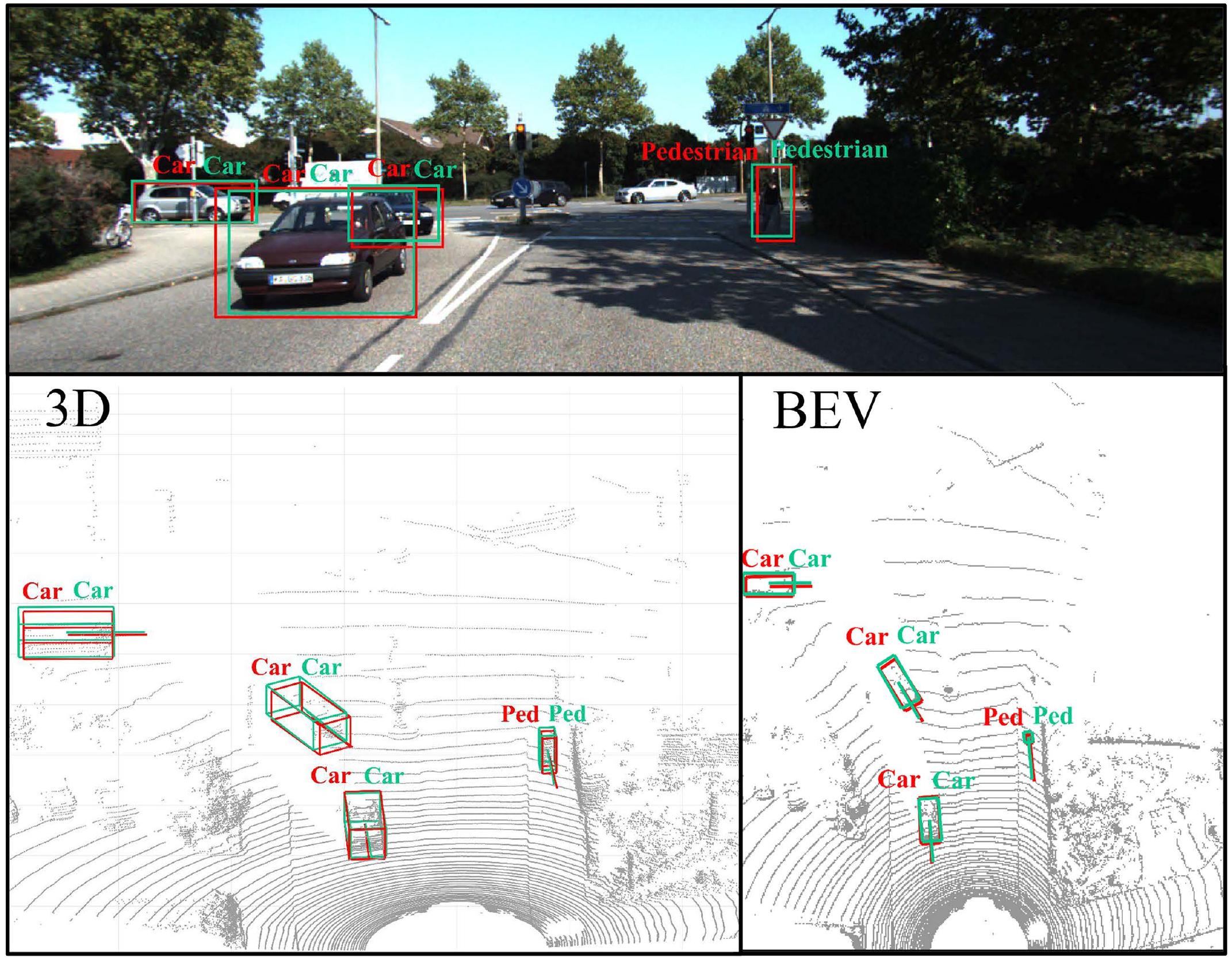}}\hspace{5pt}
	\subfloat[]{\includegraphics[scale=0.3]{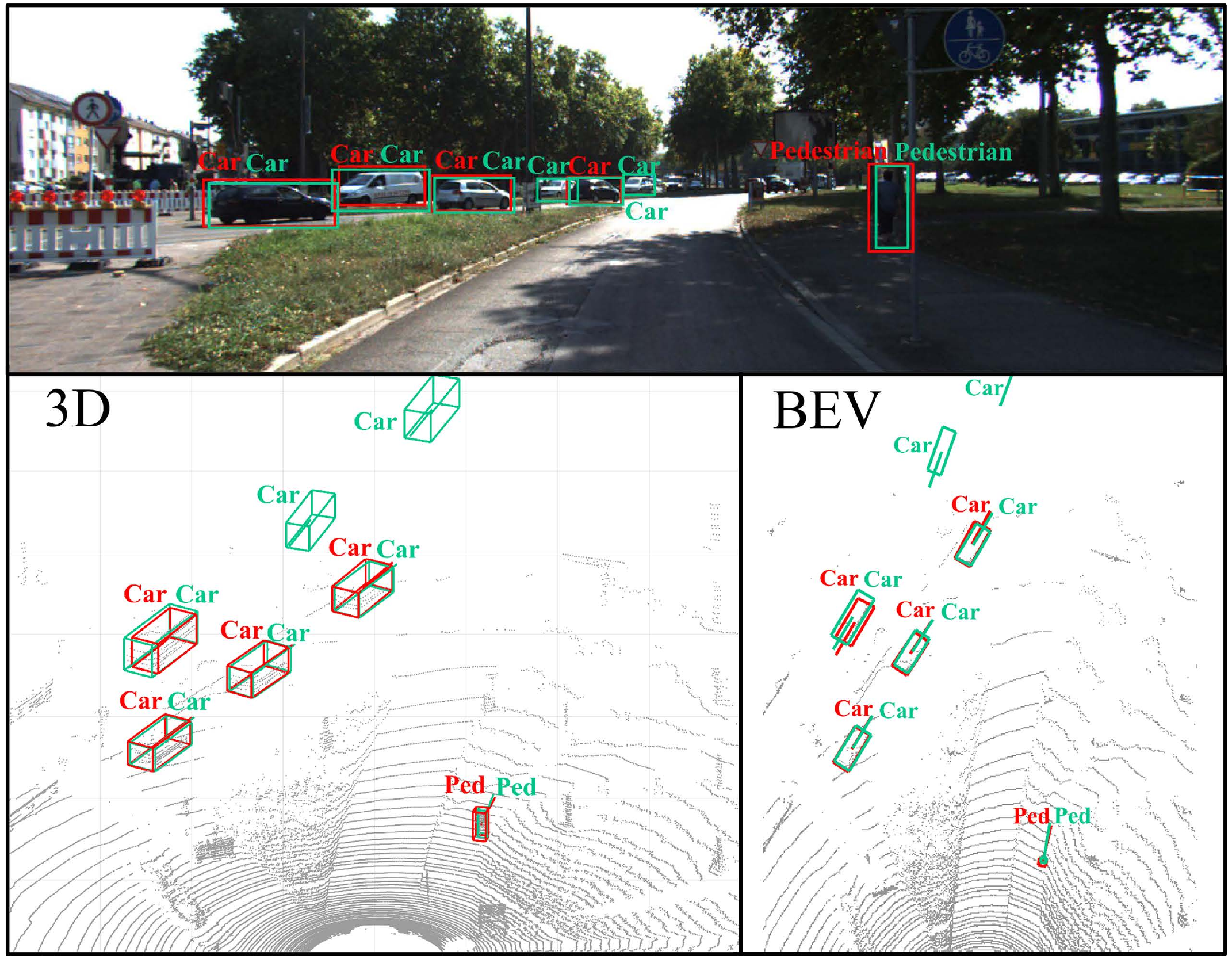}}
	\caption{\textbf{The results of our algorithm.} The red boxes represent our detection results, and the green boxes represent the groundtruth. In each subgraph, the bottom-left part shows 3D results, and the bottom-right shows the BEV results.}
	\label{fine_results}
\end{figure*} 

\begin{figure*}[!t]
	\centering
	\subfloat[]{\includegraphics[scale=0.45]{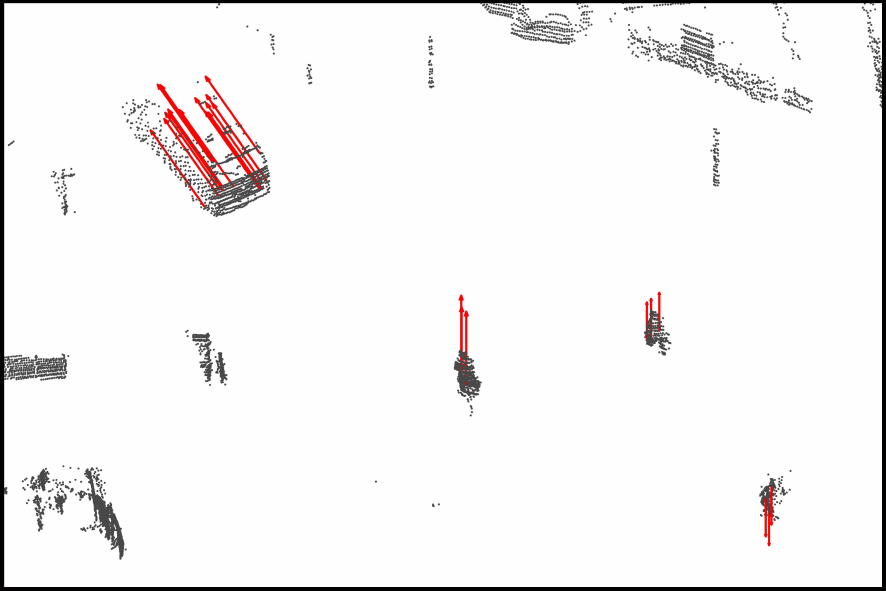}} \hspace{0.1pt}
	\subfloat[]{\includegraphics[scale=0.45]{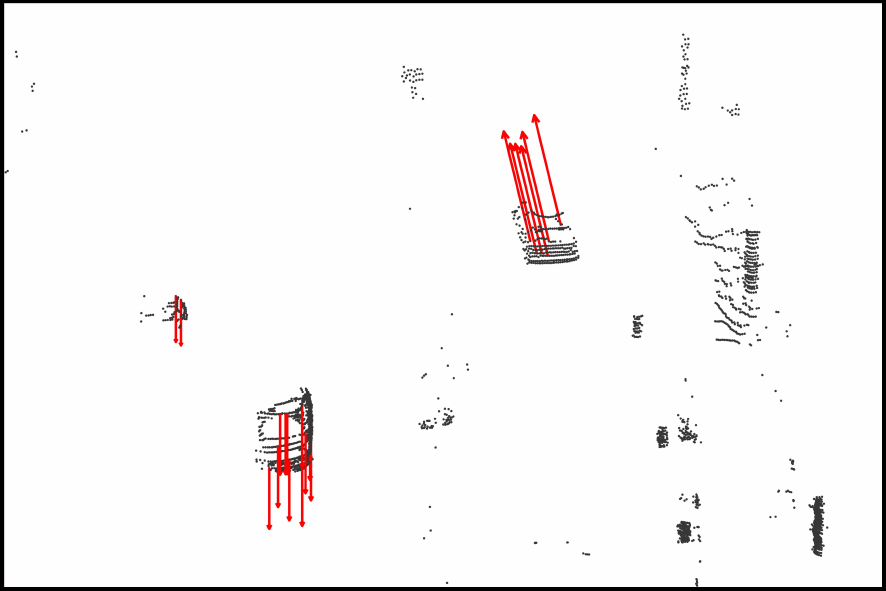}} \hspace{0.1pt}
	\subfloat[]{\includegraphics[scale=0.45]{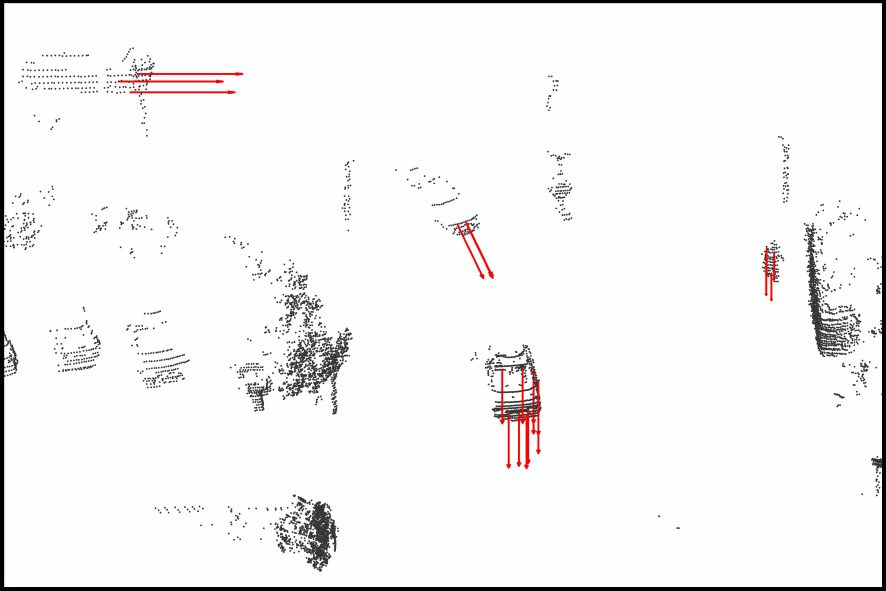}} \\
	\subfloat[]{\includegraphics[scale=0.45]{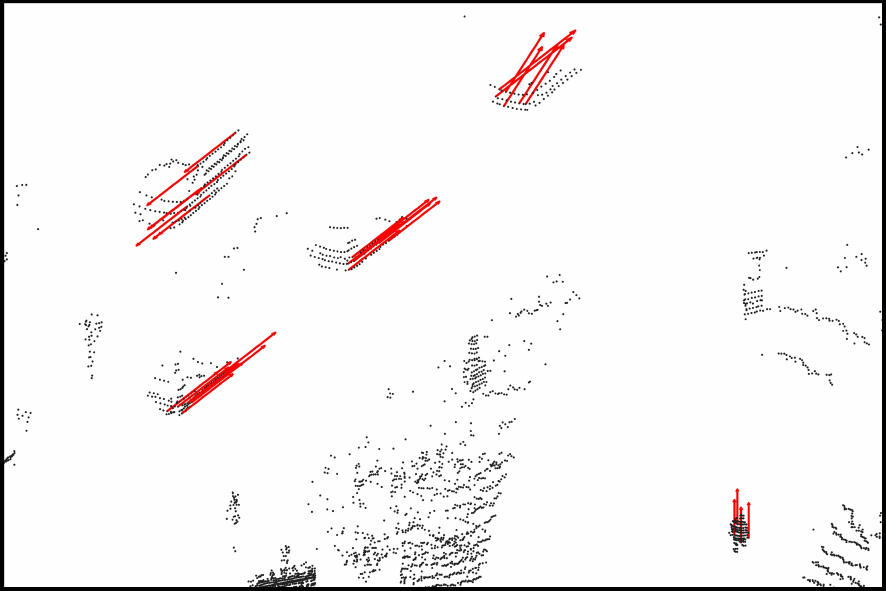}} \hspace{0.1pt}
	\subfloat[]{\includegraphics[scale=0.45]{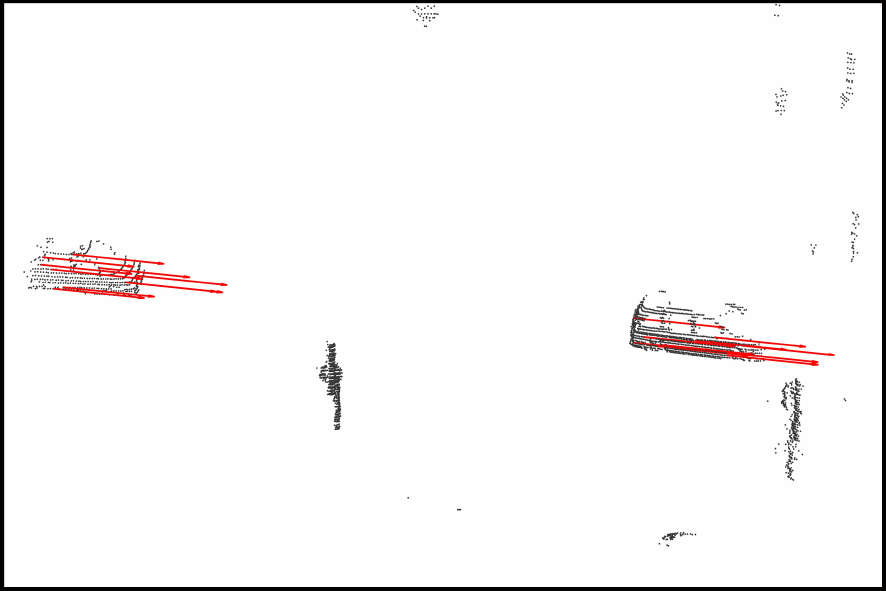}} \hspace{0.1pt}
	\subfloat[]{\includegraphics[scale=0.45]{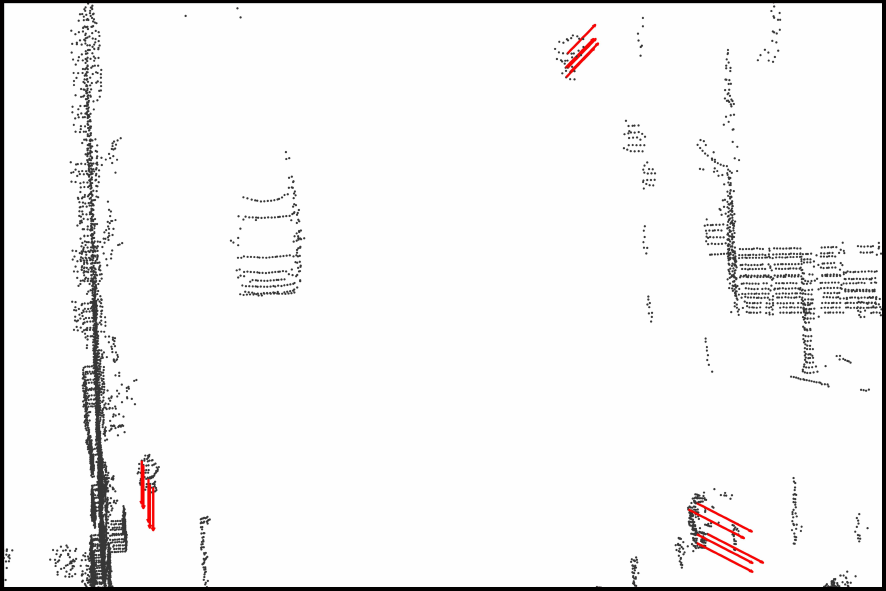}} \\
	\caption{\textbf{Lidar point flow.} These pictures show the point flow of Lidar scene, describing the movement speed and direction of each Lidar point. (a)$\sim$(d) respectively correspond to the motion scene in Fig.\ref{fine_results}.}
	\label{flow_graph}
\end{figure*}

\begin{figure*}[!t]
	\centering
	\subfloat[]{\includegraphics[scale=0.88]{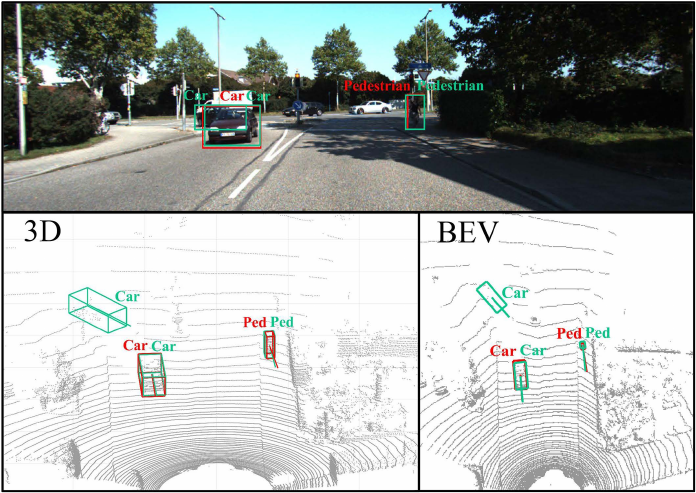}} \hspace{5pt}
	\subfloat[]{\includegraphics[scale=0.88]{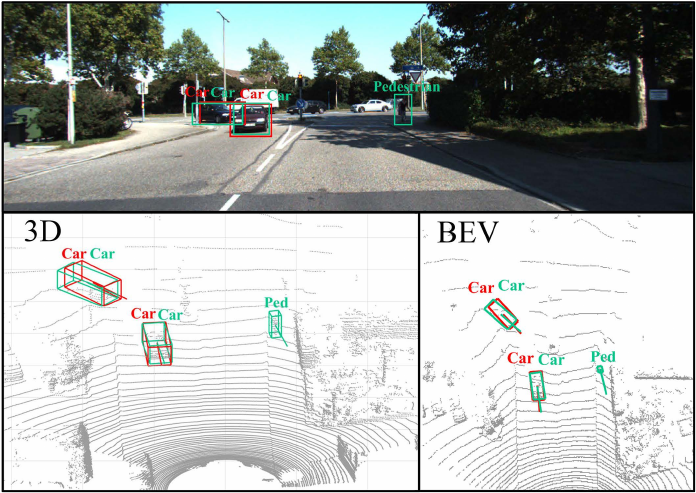}} \\
	\subfloat[]{\includegraphics[scale=0.88]{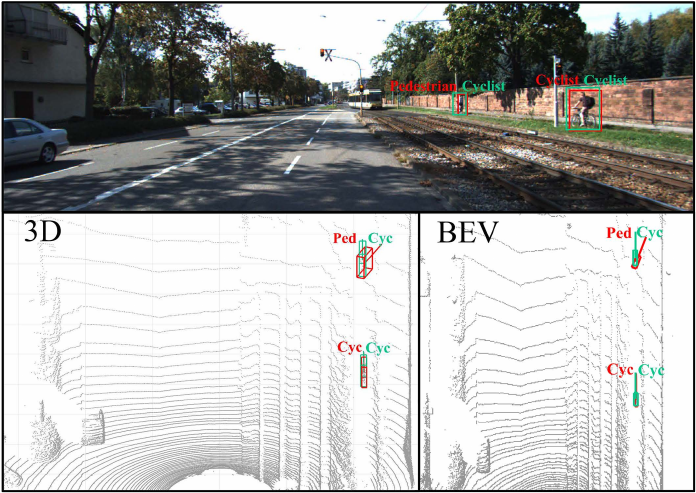}}\hspace{5pt}
	\subfloat[]{\includegraphics[scale=0.88]{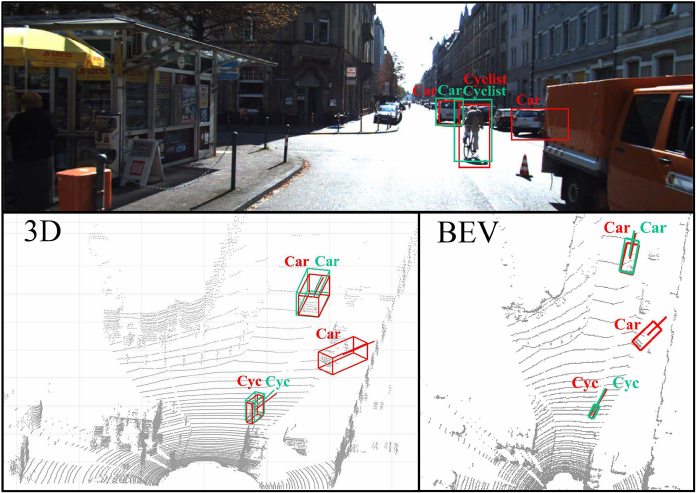}}
	\caption{\textbf{The failure cases of detection results.} (a) describes the miss due to occlusion; (b) expresses the miss because of the slow speed of pedestrian; (c) denotes that sparse point clouds cause object classification errors; (d) indicates that the static object is misdetected due to the point cloud changes caused by occlusion.}
	\label{error_results}
\end{figure*} 

The above series of experiments verify the effectiveness of the proposed method. Furthermore, Fig.\ref{fine_results}, Fig.\ref{flow_graph} and Fig.\ref{error_results} show more qualitative results. Fig.\ref{fine_results} shows some typical object detection results. From the figure, it can be seen that our detection results are almost identical to the groundtruth. Fig.\ref{flow_graph} describes the object motion flow. The results correctly reflect the motion tendency of the dynamic objects. 

Table \ref{Tab_1} shows that our algorithm achieves a recall of 84.31\% in 3D evaluation indicators. Those missing detections mostly occur in the far distance range, as shown in Fig.\ref{fine_results}(d). This is also the reason why the PR curve is cut off earlier in Fig.\ref{PR_Network}. We set the distance threshold to 30 meters and re-evaluate our algorithm. As shown in Fig.\ref{recall_distance}, it is obvious that the recall increases significantly, and the recall of detected cars is close to 1.

However, the proposed method performs not good enough for detecting small moving objects in the distance. Besides, the scene changes caused by heavy occlusion may also cause false detection of our method. Some failure cases are as shown in Fig.\ref{error_results}. Fig.11 (a) shows a missed detection scenario due to heavy occlusion. Failure caused by the sparse Lidar points often occurs at further distances (Fig.11 (c)). Because of the slow-moving speed and the small size, our method doesn't detect the walking pedestrians in Fig.11 (b). As shown in Fig. 11 (d), the static object is mistakenly detected owing to the change of the point cloud caused by occlusion.

\section{Conclusion}\label{sec:conclusion}
In this paper, we have proposed a novel bio-inspired 3D moving object detection approach. It is composed of two stages. In the first stage, the coarse-to-fine motion detector searches the motion area based on the EMD model and extract the moving point clouds to form object proposals. In the second stage, the point clouds of the proposals are segmented into foreground and background. Our network then classifies the foreground points and predicts the 3D boxes. We evaluate the proposed approach on the KITTI benchmark. The results indicate that our approach performs better than most of the comparing approaches.

In the future, we will consider integrating the multi-object tracking method to improve the proposed algorithm. Besides, based on this work, we will conduct researches on the behavior prediction of moving objects.

\section*{Acknowledgments}
This work is supported by the National Natural Science Foundation of China under Grant NO.61803380, No.61790565 and No.61825305. The authors would like to thank the laboratory and all partners.

\bibliography{mybibfile}

\end{document}